\documentclass[12pt, letterpaper]{article}
\usepackage{graphicx} 

\usepackage{blindtext}
\usepackage{multicol}

\usepackage{enumitem}
\usepackage[utf8]{inputenc}
\usepackage[T1]{fontenc}
\usepackage{mathptmx} 
\usepackage[margin=1in]{geometry}
\usepackage{amsmath}
\usepackage{float}
\usepackage{subfig}

\usepackage[natbibapa]{apacite} 

\usepackage[dvipsnames]{xcolor} 
\def\red#1{{\color{red}#1}}  

\newcommand{\nc}{\normalcolor}
\newcommand{\magenta}{\color{magenta}}

\usepackage{xifthen}
\newcommand{\figref}[3]{\ifthenelse{\isempty{#3}}
  {
    \hyperref[#2]{#1} \ref{#2}} 
  {
    \hyperref[#2]{#1 }\ref{#2} \hyperref[#2]{#3}}
   \unskip
   }

\usepackage{hyperref}
\hypersetup{
    colorlinks=true,
    linkcolor=blue,
    filecolor=magenta,      
    urlcolor=blue,
    citecolor=blue,
}

\usepackage{csvsimple}
\usepackage{longtable}
\usepackage{booktabs}

\newcommand{\f}[2]{\frac{#1}{#2}}

\usepackage{comment}
\usepackage{adjustbox}

\usepackage[labelfont=bf]{caption}
\usepackage[parfill]{parskip} 

\title{\vspace{-4em}\fontsize{26}{26}\selectfont En masse scanning and automated surfacing \\ of small objects using Micro-CT\vspace{-1em}}

\author{
Riley C. W. O'Neill$^1$\footnote{corresponding author} \\ 
\small \href{mailto:oneil571@umn.edu}{oneil571@umn.edu}\\
\small OrcidID: 0000-0001-7492-2540
\and  Katrina Yezzi-Woodley$^2$\\
\small \href{mailto:yezz0003@umn.edu}{yezz0003@umn.edu}\\
\small OrcidID: 0000-0001-7745-8069
\and Jeff Calder$^1$ \\
\small \href{mailto:jwcalder@umn.edu}{jwcalder@umn.edu}\\
\small OrcidID: 0000-0002-9829-4128
\and  Peter Olver$^1$ \\
\small\href{mailto:olver@umn.edu}{olver@umn.edu}\\
\small OrcidID: 0000-0001-6209-8777
}

\date{\small \em \vspace{-1em} $^1$School of Mathematics, University of Minnesota, Minneapolis, MN, USA\\%
    $^2$Department of Anthropology, University of Minnesota, Minneapolis, MN, USA\\%
}

\begin{document}

\maketitle
\vspace{-1em}
\begin{abstract}
\noindent Modern archaeological methods increasingly utilize 3D virtual representations of objects, computationally intensive analyses, high resolution scanning, large datasets, and machine learning. With higher resolution scans, challenges surrounding computational power, memory, and file storage quickly arise. Processing and analyzing high resolution scans often requires memory-intensive workflows, which are infeasible for most computers and increasingly necessitate the use of super-computers or innovative methods for processing on standard computers. Here we introduce a novel protocol for en-masse micro-CT scanning of small objects with a {\em mostly-automated} processing workflow that functions in memory-limited settings. We scanned 1,112 animal bone fragments using just 10 micro-CT scans, which were post-processed into individual PLY files. Notably, our methods can be applied to any object (with discernible density from the packaging material) making this method applicable to a variety of inquiries and fields including paleontology, geology, electrical engineering, and materials science. Further, our methods may immediately be adopted by scanning institutes to pool customer orders together and offer more affordable scanning. The work presented herein is part of a larger program facilitated by the international and multi-disciplinary research consortium known as Anthropological and Mathematical Analysis of Archaeological and Zooarchaeological Evidence (AMAAZE). AMAAZE unites experts in anthropology, mathematics, and computer science to develop new methods for mass-scale virtual archaeological research. Overall, our new scanning method and processing workflows lay the groundwork and set the standard for future mass-scale, high resolution scanning studies. 
{\bf Statements and Declarations}: \\ Competing interests: the authors declare no competing interests. 
\end{abstract}

\newpage

\section{Introduction}
3D modeling via high-resolution scanning is becoming prolific within archaeological research and, more broadly, anthropological research. This has enabled the development of new data extraction techniques and the application of computationally intensive analyses and machine learning \citep[e.g.][]{yezzi2024using, calder2022use, mcpherron2022machine, orellana2021proof, yezzi2021virtual, yezzi2022batch, spyrou2022digital, hostettler20243, schunk2023enhancing, muller2024automatic, wyatt2022after, goldner2022practical, goldner2023styrostone, hipsley2020high}.

Scanning multiple objects simultaneously using computed tomography (CT) scanners previously required tremendous computing power, memory, and time-consuming manual segmentation done by humans. \cite{hipsley2020high} developed a batched micro-CT scanning protocol by arranging straws filled with very small objects  in a 50 ml tube. Using their method, 80-253 small objects (<1cm in length) can be scanned simultaneously. Yet, all post processing is manual and was done in proprietary software - manually segmenting, naming, and saving objects with the corresponding identifier using hand-drawn, labeled diagrams for each scan \citep{hipsley2020high}. They note that as the packing number increases, the packaging tend to shift and not align with the diagram. Concurrent developments by \cite{goldner2022practical} and \cite{yezzi2022batch} were among the first protocols developed within archaeology for mass scale micro-CT and CT scanning of larger objects. \cite{goldner2022practical} introduced a  mass-scale micro-CT scanning and surfacing protocol for lithics. Two sides of a styrofoam block are manually carved out to fit each each object, recording the side, row, column, and identifier for each object. After scanning, the scan is cleaned, surfaced, segmented into each side, segmented into each row, segmented into each object, then saved after entering the correct identifier for the file name \citep{goldner2023styrostone}. All of this postprocessing is again manually done by a human and, in part, uses proprietary softwares. While \citet{goldner2023styrostone}'s advancement makes the scanning process itself much more efficient, the tremendous amount of manual human intervention required in post-processing is very time-consuming, repetitive, and exhausting. The initial surfacing and cleaning here likely requires a computer with large memory (RAM) to be able to operate on the entire scan simultaneously, making it infeasible to run on most consumer-grade computers. In contrast, \cite{yezzi2022batch} introduced an automated workflow for batched medical CT scanning. This utilizes an automated post-processing workflow designed to scan and surface objects packed in a row with space separating them. Maintaining a CSV file describing the object identifiers (e.g. accession number) and scan layout, the row of objects can be automatically segmented into individual surfaces with the correct identifiers \citep{yezzi2022batch}. However, medical CT scanning doesn't offer the resolution of micro-CT, and this doesn't allow the most efficient usage of the scanning bed. Further, both of these packing methods are not very reusable. Naturally, this incentives similar, mostly-automated, reusable, and user-friendly workflows for batched micro-CT scanning of several objects. Indeed, the continued development and extension of such high-throughput scanning methods is paramount to the implementation and innovation of new data extraction and processing pipelines from large 3D mesh datasets. 

Here we introduce a new, fully-replicable protocol for mass scale scanning of small objects using micro-CT. Packaging specimens for scanning is low-cost, efficient, and user-friendly. In contrast to previous en-mass micro-CT protocols, the segmentation workflow (shown in \figref{Figure}{fig:workflow}{}) requires minimal human intervention to process the scans (only a few minutes per scan as opposed to a few hours). Our method overcomes numerous memory constraints so it runs on any computer with at least 16GB of RAM, removing the need for expensive computational resources. Additionally, our method demonstrates drastic cost and time savings. This method has the potential to revolutionize the scanning of large collections - not just in archaeology, but beyond. In addition to use by other mass-scale high-resolution scanning studies to scan many different materials, our protocol may be adopted by scanning centers to drastically lower the cost of high-resolution scanning by pooling several customers' orders together, making this method effective for any discipline or project where numerous objects can be scanned by micro-CT. Furthermore, our processing code is open access and freely available via the \href{https://amaaze.umn.edu}{AMAAZE website} and \href{https://github.com/oneil571/AMAAZE-MCT-Processing}{Github}, and we will begin to make our models available on an open-access database by the end of 2024.

\begin{figure}
    \centering
    \includegraphics[width=0.9\linewidth, clip= true, trim = 0 0 70 0]{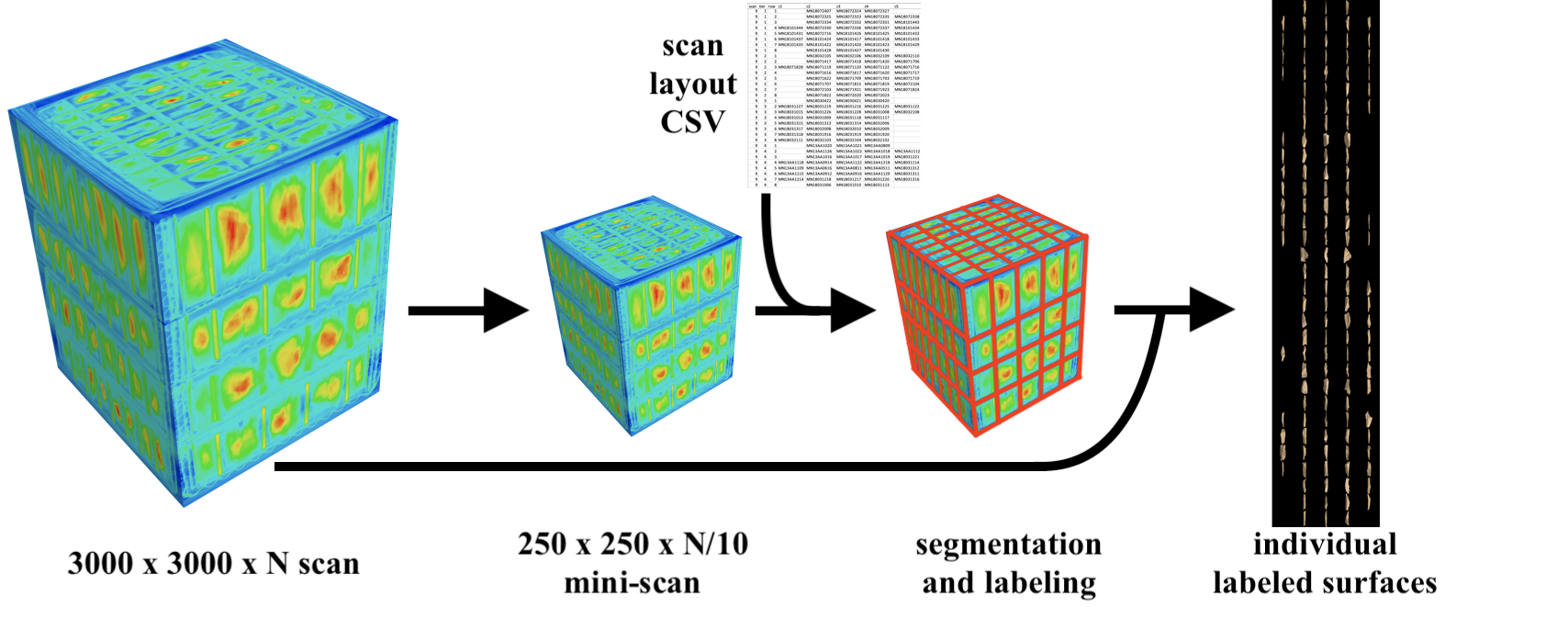}
    \caption{Overview of our processing workflow.}
    \label{fig:workflow}
\end{figure}

\section{Materials and Methods}

\subsection{Sample}

Bone fragments (N = 1,151) were acquired from a collection of experimentally broken appendicular long bones from various ungulates including cow (\textit{Bos taurus}), horse  (\textit{Equus caballus}), sheep (\textit{Ovis aries}), elk (\textit{Cervus canadensis}), and deer (\textit{Odocoileus virginianus}).  Some bones were broken by people (mostly graduate and undergraduate students) using stone tools to emulate a way early hominins may have broken bones using a hammerstone and anvil or simply by striking the bone against the rock that was used for the anvil. Others were fed to spotted hyenas (\textit{Crocuta crocuta}) at either the Milwaukee County Zoo or the Irvine Park Zoo, both in Wisconsin. The subsequent fragments were submerged in water and cleaned at low heat using crockpots then set out to air dry prior to scanning \cite[See][for a more detailed descriptions of the protocols used for acquiring the physical bone fragments.]{coil2020comparisons, yezzi2024using, yezzi2021virtual}. The fragments used in this study had a maximum length of 7.27 cm, maximum width of 2.55 cm and height of 1.46 cm (smallest axis of the overall bounding box). The complete list of specimens can be found in \figref{Table}{tab:specimens}{} in Supplemental Information. 

\subsection{Packing and Imaging}

A total of 1,153 bone fragments were scanned. Fragments were packed into corrugated AVIDITI brand cardboard boxes (9"L x 4"W x 4"H). A package of 25 boxes costs $\sim$ $\$$27.00 USD through Amazon. Each fragment was placed into an individual cell created by interlocking plastic drawer dividers. A package of 40 Flytianmy brand drawer dividers can be purchased from Amazaon for $\$$22.99 USD. Each interlocking piece is $2.76$ inches tall and $12.6$ inches long but can be cut into shorter pieces (see \figref{Figure}{fig:pack}{A}). The height of each divider dictated the allowable length of the fragment. The way in which the dividers were interlocked dictated the allowable width and depth of each fragment. One interlocking grid of dividers constituted a tier. One to four tiers were placed in each box, stacked one on top of the other and separated by a thin piece of cardboard (see \figref{Figure}{fig:pack}{B}). Each tier contained up to 40 fragments and each box contained between 30 and 228 fragments, as shown in full in \figref{Table}{tab:scan_nums}{}. A total of 10 boxes were individually scanned. Handwritten labels were placed on the exterior of each box to ensure that they were properly oriented on the scanning bed.

\begin{figure}[H]
    \centering
    \includegraphics[width=\textwidth]{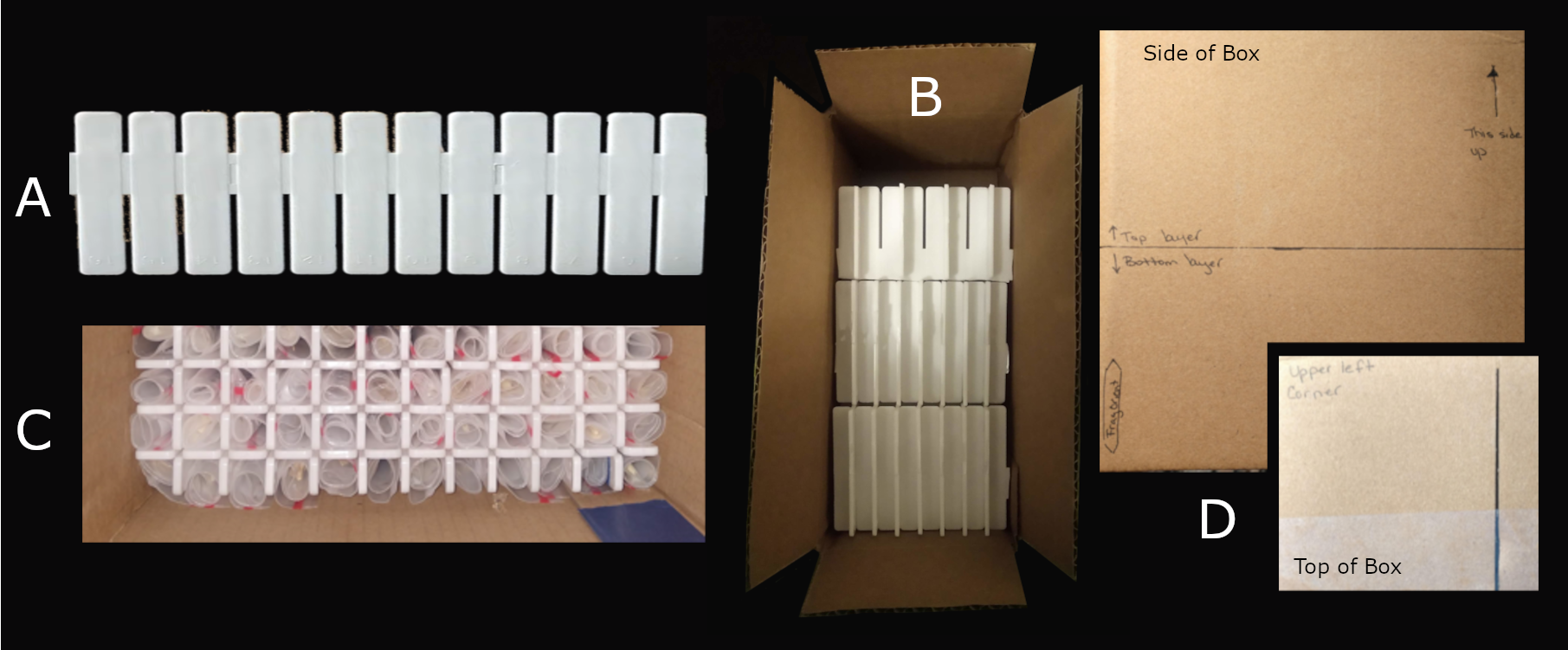}
    \caption{{\bf A} is the plastic divider. {\bf B} is a side view of the scan as it would be placed on the scan bed. Note that the topmost tier is shorter than the other two tiers, demonstrating one way the sizing can be altered to accommodate different sized objects.  The extra space at the top of the package was to ensure everything would be captured during scanning. {\bf C} is a bird's-eye-view of a portion of tier. You can see how each fragment is rolled in its plastic bag and inserted into one of the cells. The cell lengths and widths may be varied by adjusting the divider placement and skipping slots. {\bf D} shows labeling on the side and top of the box so that it was clear how the box needed to be set on the scan bed and how it related to the CSV.}
    \label{fig:pack}
\end{figure}

The fragments are small (max length 7.27 cm) and thus are curated separately in individual plastic bags. In most cases, the fragments remained in their collections bag which was rolled up and inserted into the cell (see \figref{Figure}{fig:pack}{C}). In some cases this would create too much bulk and the fragment had to be removed from its bag. In these cases, extra care was taken to ensure that the fragment did not become permanently detached from its label. This was done either through careful note taking and/or including a handwritten label in the cell. The outside of the box was labeled with the appropriate orientation (see \figref{Figure}{fig:pack}{D}). After the first scan, the box size was reduced to the size provided herein and, in some cases, extra space was left at either the top or the bottom of the box as can be seen in \figref{Figure}{fig:pack}{B}. The purpose of this was to ensure all objects would be within the field of view and not cut off in the scan.

\begin{table}
\begin{center}

\begin{tabular}{|c|c|}
     \hline
     Scan & Framents (N)\\
     \hline
     \hline
       1  & 228 \\ 
       2 & 30 \\
       3 & 86 \\
       4 & 134 \\
       5  & 134  \\
       6 & 102 \\
       7 & 109 \\
       8 & 101 \\
       9 & 133 \\
       10 & 96 \\
       \hline
\end{tabular}
\caption{Number of objects (bone fragments) in each scan.}
\label{tab:scan_nums}
\end{center}
\end{table}

During the packing process, the arrangement of the objects within each scan was documented in a CSV file we refer to as the "scan layout." Our post processing algorithm crossreferences this essential document to retrieve the object identifiers to properly name the 3D model file of each scanned fragment. \figref{Figure}{fig:csv}{} shows an except of this CSV file for a 4-tiered scanning package. The first column denotes the scan number (so as to differentiate it from others within the same CSV), and the second column the tier. If looking at rows of the same tier number, it gives an overhead image of the arrangement of fragments within the tier - this is essential to recovering the identities of the constituent fragments. While each tier could hold up to 40 fragments, at least 2 cells in each tier were left empty to form a unique pattern for each scanning package. This was done to easily recover the scan orientation, rule out mirroring, act as a fail-safe if the scan's identity was somehow lost after scanning, and align the CSV layout and the scan for the segmentation algorithm. 

Each box of fragments was scanned using a North Star Imaging X5000 micro-CT scanner with a twin head 225 kV x-ray source and a 3073 x 3889 pixel Dexela area detector, housed at the University of Minnesota's ESci XRCT lab. Each scan took between 1 and 1.5 unsupervised hours.

\begin{figure}[H]
    \centering
    \includegraphics[width=.9\textwidth]{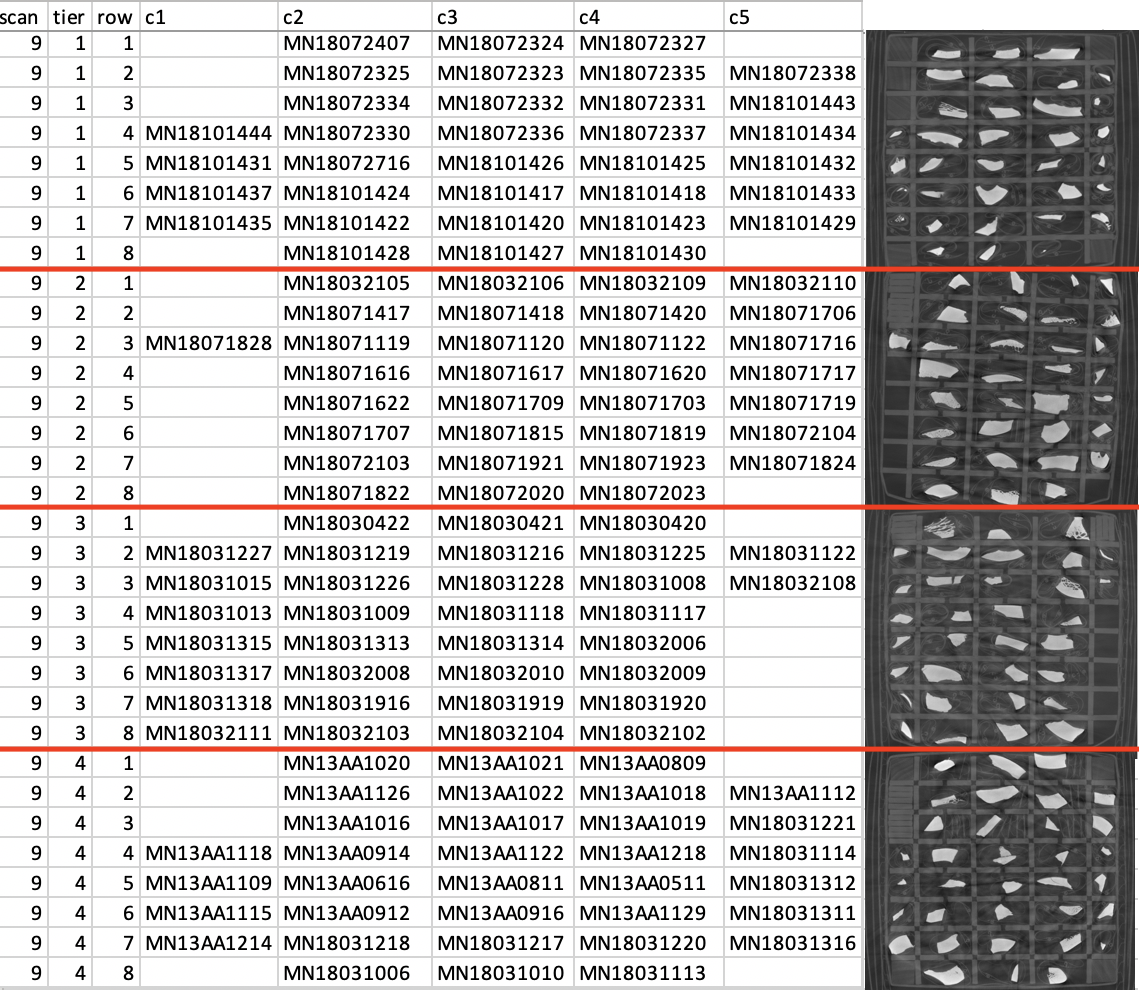}\\
    \caption{Left: example "scan layout" CSV file for a 4-tier micro-CT scan. Note red lines were added to emphasize different tiers. Each row contains a scan identifier, tier number, row number, and the identifiers corresponding to the objects in that strip of the scan (where present). Right: slices from the corresponding tiers in the scan. }
    \label{fig:csv}
\end{figure}

\begin{figure}[H]
    \centering
    \includegraphics[width=.9\textwidth, clip =true, trim= 0 25 0 28]{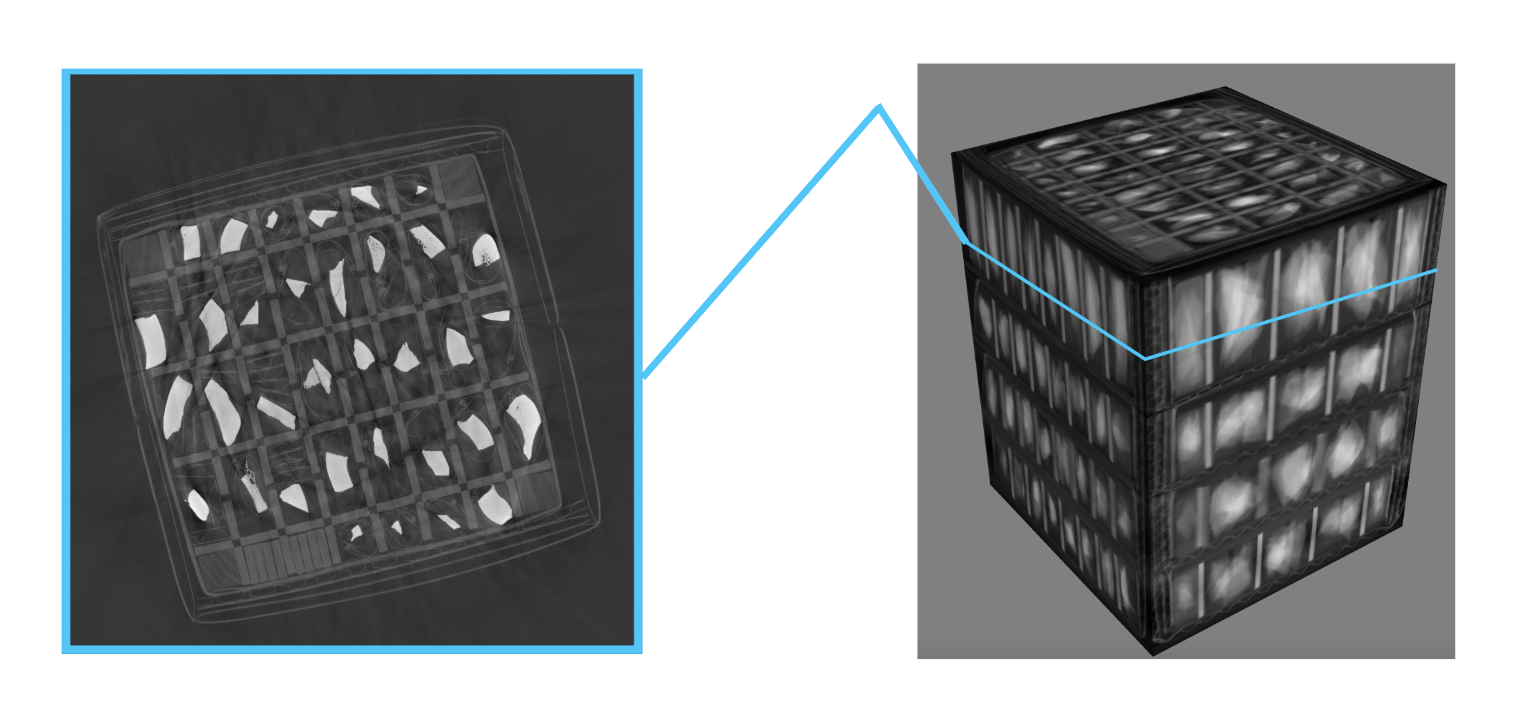}
    \caption{left: a $z$-slice TIFF image of the scan. Right: visualization of the entire scan.}
    \label{fig:zslice}
\end{figure}

\subsection{Scan Processing}

As each micro-CT scan is 70+ GB, some unique challenges arise in processing the scans. For one, an entire scan's volume cannot be loaded into a conventional computer's memory. Even the extracted sub-volumes containing each object are several gigabytes. Thus, all scans were stored on a 4 TB drive. To mitigate the memory-usage challenges of the large scans, we developed a semi-automated processing scheme that builds on the work of \cite{yezzi2022batch}. It consists of 5 key steps:
\begin{enumerate}
\item {\bf Initial rotation and cropping}, to align the scan with the scan layout CSV. 
\item {\bf Subsampling}, to recover a mini-scan that fits in memory for:
\item {\bf Bounding box identification} for each fragment cell (on subsampled volume).
\item {\bf Sub-volume extraction} for each fragment cell (on full scan).
\item {\bf Surfacing} of each fragment via marching cubes and cleaning. 
\end{enumerate}

Note that our scans were stored as TIFF $z$-slice images (\figref{Figure}{fig:zslice}{}) on a remote server. Visualizations and bounding-box identification (steps 1 and 3) were done on a 2017 Apple MacBook Pro laptop (250 GB storage, 16 GB memory), while all other operations were done on a System76 machine with a 3960X processor, 48 CPUs, 256 GB Quad channel DDR4 @ 3200 MHz (8x32GB), and 6TB of storage. While all operations can be performed a machine with 16GB of memory, drastic speedups can be obtained with {\em parallelization}, where processes are divided into smaller tasks that can be run simultaneously over several CPU cores (as opposed to running everything on a single CPU). Most consumer computers do have several CPUs today, but a good-sized memory is required for parallelization to leverage as many cores as possible here. For the most time consuming processes, i.e. sub-volume extraction, surfacing, and cleaning, parallelization contributes to a drastic speedup. Intermediary data and the ouputs of steps 1, 2, and 3 were exchanged between the two devices using SSH (SecureSHell)\footnote{Note that portions of our Python code are designed to use SSH and will have logins and passwords redacted prior to publication so as to ensure the security of our server; functionality for portable drives and local files was not the original use-case but has been added for publication.}. 

\subsubsection{Initial Rotation and Subsampling (Steps 1 and 2)}
\begin{figure}[H]
    \centering \includegraphics[width = .95\textwidth]{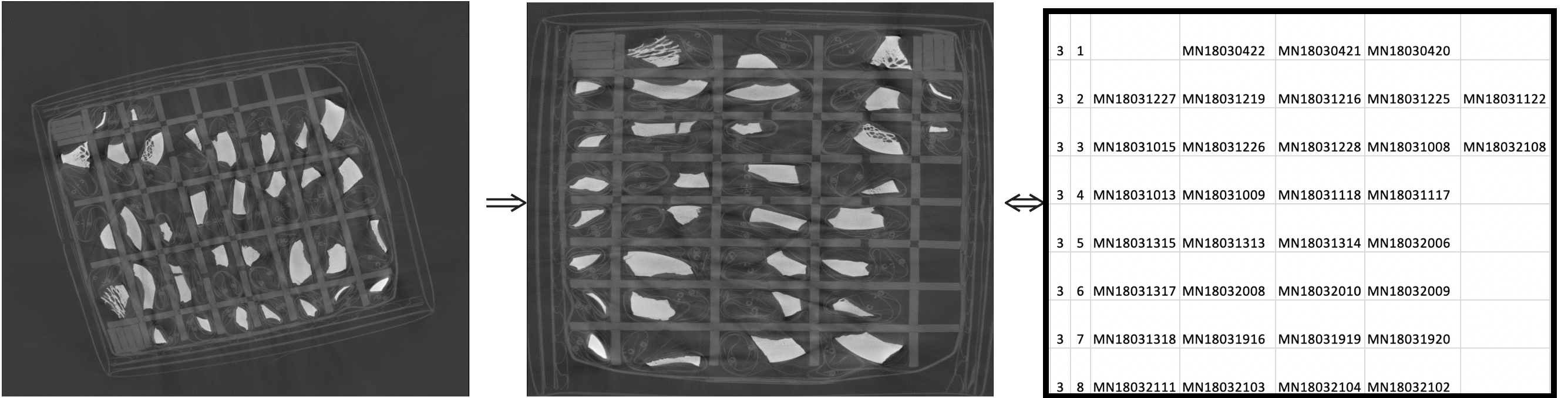}
    \caption{Left: original TIFF image. Middle: cropped and rotated image. Right: CSV layout.}
    \label{fig:im2csv}
\end{figure}

While the scanning packages were placed on the micro-CT scanning bed following the "this side up" labels (\figref{Figure}{fig:pack}{D}), no additional alignment was done - they took random positions within the scanning bed, especially regarding rotation in the $x$-$y$x plane. Thus, 2-6 slices from each scan were manually inspected to align the vertical view with the scan layout CSV via rotation and cropping (the latter simply to remove empty space). As described earlier, fragments were arranged so as to ensure the orientation could be always recovered from their arrangement alone, i.e. cells were selectively left empty to preclude rotational and reflective symmetry. The angle and row and column cropping ranges (start index, stop index) were identified by trial and error; an example is shown in \figref{Figure}{fig:im2csv}{}. This took only a few minutes for each scan. The 5 identified values were then stored in a text file for each scan. While the original scans could have been cropped, rotated, and saved to reduce data-storage, this was ultimately not done to ensure data fidelity and avoid corruption or other adverse possibilities.
 
Each TIFF $z$-slice of the original micro-CT scan was 2952 by 2971 pixels. Rotating and cropping the scan to just the scanning package gave about 2500 by 2500 pixel images for all scans except \#1. The scans varied in height ($z$-direction) - the tallest had $h = 3813$ images. The rotated and cropped scans were subsampled by resizing the $z$-slices to 225 by 225 images and then averaging together batches of 10 slices (or however many remained at the top of the scan) for a $255 \times 225 \times \text{ceil}(h/10)$. This subsampled grid could be then completely loaded into standard machine memory for bounding box identification.

\subsubsection{Bounding Box Identification (Step 3)}
Bounding box identification consists of 5 phases: (1) identification of object and divider voxel ranges and thresholding; (2) vertical ($z$) segmentation into individual tiers; (3) divider detection and automatic tier rotation; (4) automatic rotation; and (5) grid segmentation. 

\textbf{Phase 1: Identification of voxel value ranges and thresholding.} \label{sec:thresh}
While medical CT scanners are finely calibrated so the voxel values of an object directly correspond to its Hounsfeld units, micro-CT scanners are not. Quite contrarily, the voxel values of micro-CT scans are extremely sensative to a variety of scanner and processing settings: KV MicroAmp, table position, detector position, beam hardening, resampling, etc. This means that the voxel values for objects of the same density can vary {\em wildly} between scans - within our own data, bones started at 17000 in one scan and 38000 for another, with many other values in between. Despite these variations, all the voxel histograms "looked the same" (up to translation), resembling a triple Gaussian. This is shown in in \figref{Figure}{fig:isolevel}{} - from left to right: one large peak for air/background, a smaller peak for the plastic dividers, and a very low and wide peak for the objects (bones). 

Since the placement of the plastic dividers and cardboard delineate the regions to extract and correspondence to the layout CSV, this is exactly what our algorithm uses for segmentation. Thus, it is essential to identify the start and stop of the divider range $(a_{divider}, b_{divider})$. Surfacing later requires the extent of the object of interest range as well $(a_{object}, b_{object})$; we set $b_{object} = \infty$ since nothing else was denser in the scan. These  ranges are shown in \figref{Figure}{fig:isolevel}{} (right). Several options were considered for automatic identification of these three values (e.g. triple gaussian fitting, find-peaks), these proved to be sensative to noise or initialization and struggled to identify the bone range. Instead, manual identification of $a_{divider}, b_{divider},$ and $a_{object}$ took just a few seconds per scan via a guided-user-interface.

\begin{figure}[H]
    \centering
    \includegraphics[width=.95\textwidth, clip = true, trim = 1 1 1 1]{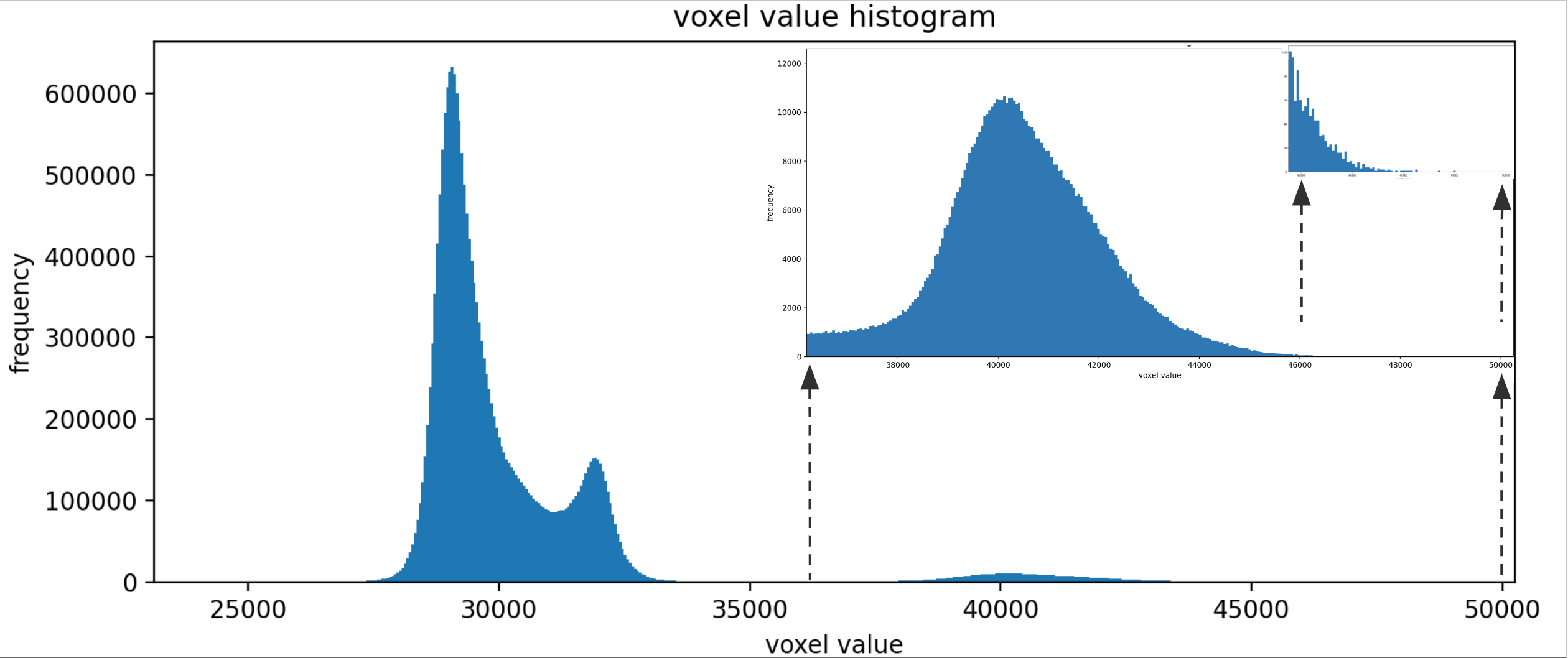} \\ 
    \includegraphics[width=.95\textwidth, clip = true, trim = 1 1 1 1]{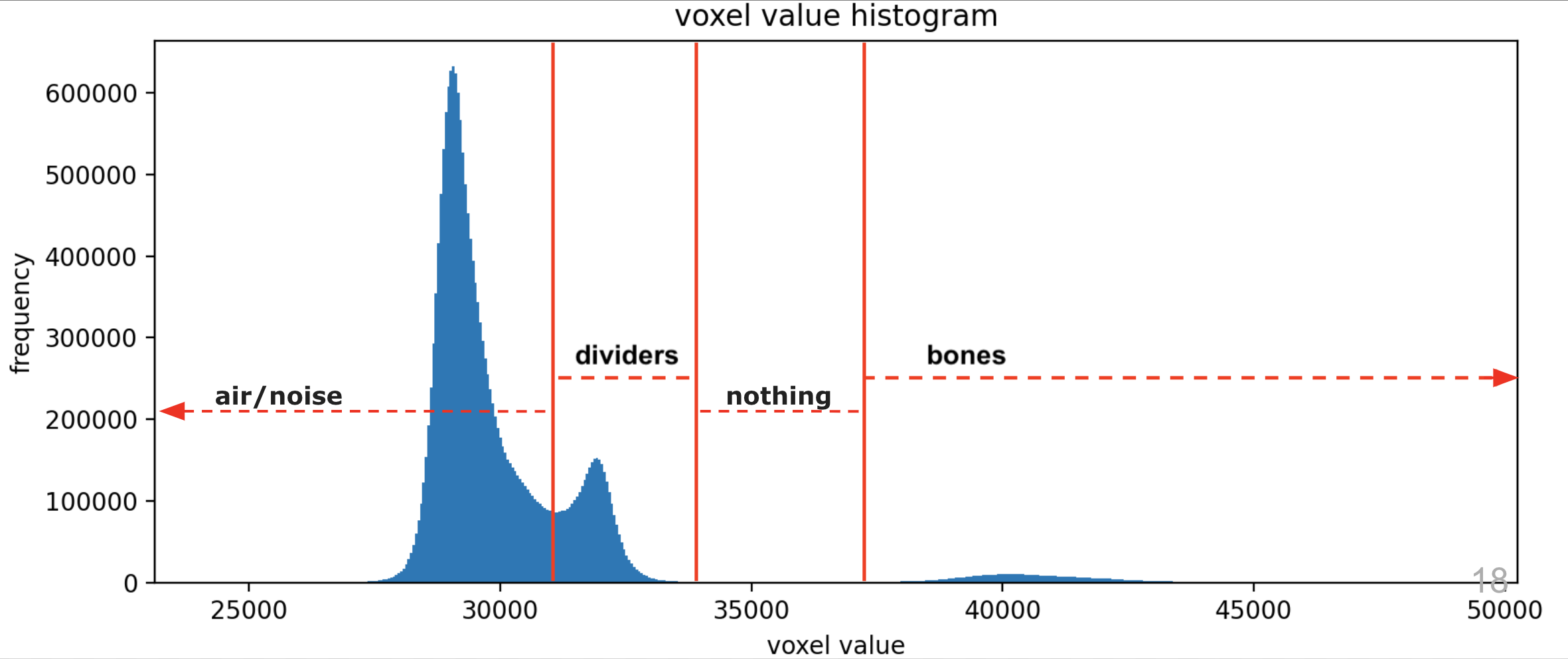}
    \caption{Left: density histogram for total subsampled volume (500 bins). The largest peak corresponds to background (air), the middle peak the cardboard, and the low right peak the bones. Right: 3 user selected threshold values to identify the cardboard range and start of the bone range.}
    \label{fig:isolevel}
\end{figure}

{\bf Phase 2: Vertical Segmentation}
Each tier consists of plastic inserts separating bones, and cardboard separates the tiers. Bone has a higher density than cardboard, so the average pixel value for a $z$-slice containing only cardboard is lower than that of a tier with plastic and bones. Let $D_k$ be the average pixel value over the $k$th $z$-image in the $225 \times 225 \times N$ subsampled volume. Peaks in $-D_k$ then correspond to the tier boundaries. The layout CSV file says how many tiers are in each scan, so we know how many peaks to look for to split the scan. We used an automated peak detection interface (minimum width = 10) with user verification and optional input to ratify or modify the detected peaks for splitting each scan into individual tiers. This is shown in \figref{Figure}{fig:zseg}{}.

\begin{figure}[H]
    \centering
    \includegraphics[width=.7\textwidth, clip=true, trim=4 5 4 0]{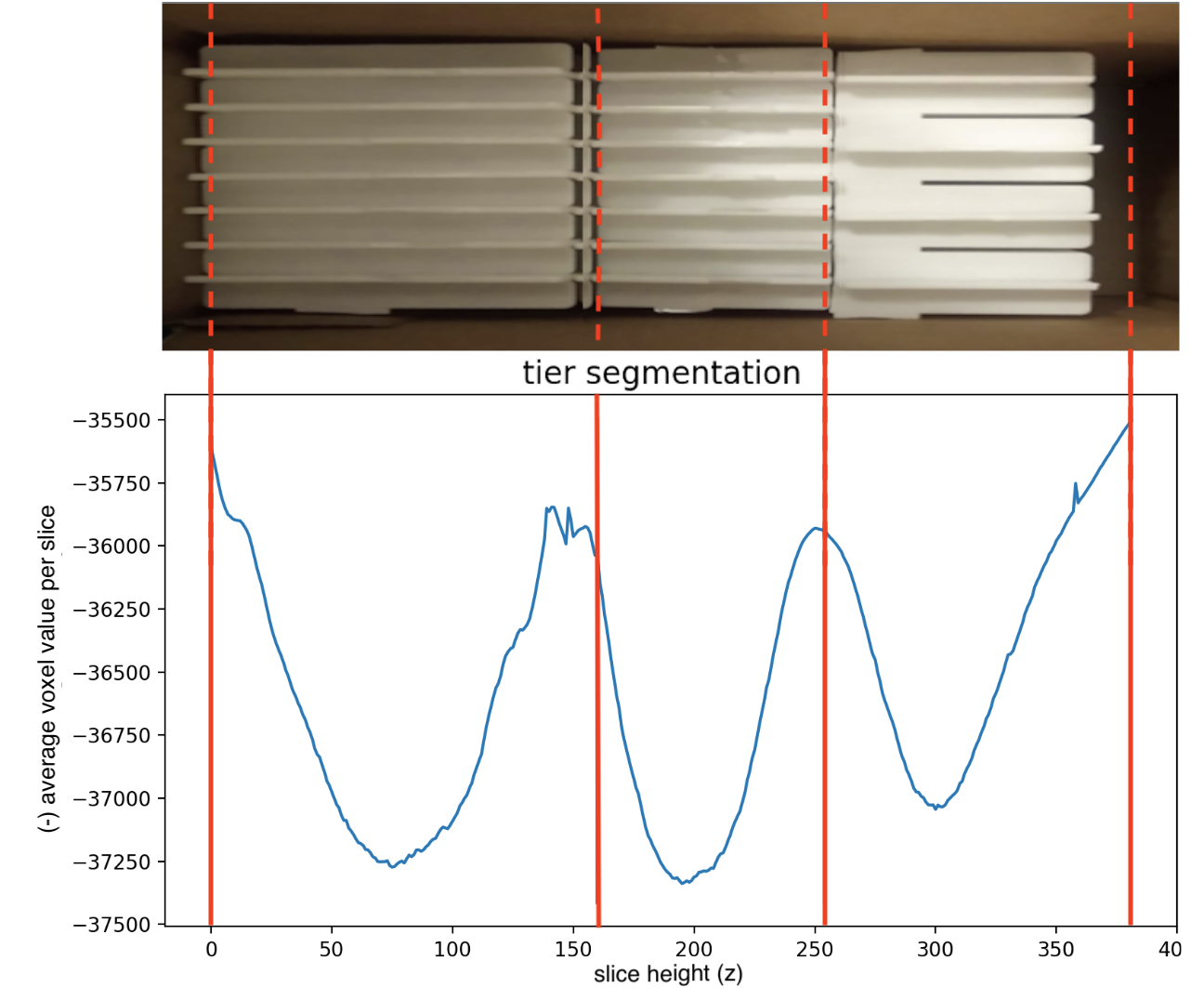}
    \caption{Top: Dividers, side view. Left to right is higher in the scan. 
    Bottom: negative average density for each tiff-image plotted against slice height subsampled volume. Peaks (corresponding to lower densities) denote boundaries between tiers - i.e. how to split the scan, while valleys denote regions with denser material (i.e. bones).  Right: segmentation into 4 tiers (user verified). I would make this look like a single figure, trim left margin of photo and extend red lines to arrows. This will mess with the tier segmentation label perhaps.}
    \label{fig:zseg}
\end{figure}

{\bf Phase 3: Divider detection}
After obtaining the user-input bounds for the divider and object ranges as described in \figref{Section}{sec:thresh}{}, we leverage these ranges to robustly extract the dividers from the tier volume. Let $V$ denote the $225\times 225 \times H_z$ tier volume, with $V_{ijk}$ a single value and $V_{ij:} = [V_{ij1}, ..., V_{ijH_z}]$ a vertical segment. We want to extract a $225\times 225$ image denoting where the dividers are. A simple yet effective way is to count how many pixels of $V_{ij:}$ are in the divider/cardboard range and subtract how many pixels in $V_{ij:}$ are in the object range, i.e.:
$$I_{ij} = \#\{ k : a_{divider} \leq V_{ijk} \leq b_{divider}\} - \#\{ k : a_{object} \leq V_{ijk} \leq b_{object}\}$$
This is quite intuitive: the divider boundaries should not sit under/over bones, so very high values are very likely dividers boundaries. Thresholding this image to values within $75\%$ of the maximum value gives very-sure dividers:
$$B_{ij} = \begin{cases} I_{ij}, & \text{if } I_{ij}>0.75 \cdot \max_{mn}(I_{mn}) \\ 0, & \text{otherwise.}
\end{cases}$$
$B$ is ultimately used for automatic rotation and chopping the tier. An example is shown in \figref{ Figure}{fig:div}{}.

\begin{figure}[H]
    \centering
    \includegraphics[width=.5\textwidth, clip=true, trim = 0 40 0 35]{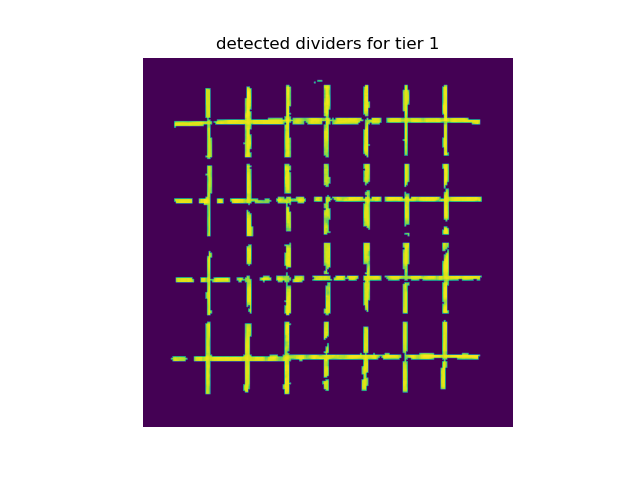}
    \caption{Detected cardboard dividers for tier 1.}
    \label{fig:div}
\end{figure}

{\bf Phase 4: Automatic Rotation of Tiers}

Once the subsampled volume has been chopped into individual tiers, it needs to be segmented into individual bounding boxes. Since we used drawer dividers stacked atop one another, the tiers do not always align squarely with one another: in some instances they slightly twist, so we developed a method to automatically rotate them (by small angles) so identifying the bounding boxes becomes very simple. 

The automatic rotation works on the single divider image per each tier ($B$, as constructed above). It seeks to maximize the horizontal/vertical detail to diagonal detail ratio. Define
$$V = \begin{bmatrix} 1 & -1 \\ 1 & -1\end{bmatrix}; \quad H = \begin{bmatrix} 1 & 1 \\ -1 & -1\end{bmatrix}; \quad D = \begin{bmatrix} 1 & -1 \\ -1 & 1\end{bmatrix}$$
and $R_a(I) $ the image rotated $a$ degrees using bilinear interpretation. We want to maximize
$$J(a) = \f{\sum_{i,j} (V*R_a(B))_{ij} + (H*R_a(I)_{ij}}{\sum_{i,j} (D*R_a(I))_{ij}}$$
over a small interval of angles, e.g. $ a \in [-10,10]$, which was done via sampling an evenly spaced partition in increments of $1/10$. An illustration of this completely automatic process is shown in \figref{Figure}{fig:auto_rot}{}.

The method works very well for the crisp divider images ($B$) for each tier. We originally considered applying it to the $z$ average of voxel values for each tier, but this was far less robust due to the presence of high-magnitude non-grid objects. In theory, $J$ should be smooth in $a$ and resemble a sinusoid. Experimentally, the numerical implementation of the bilinear interpretation gave some noise resembling Gibbs phenomenon about $a=0$, irrespective of attempts to counteract the the issue. We resolved to smooth $J$ and perform a quadratic fit to identify the maximum, which experimentally worked well.

\begin{figure}[H]
    \centering
    \includegraphics[width=.95\textwidth]{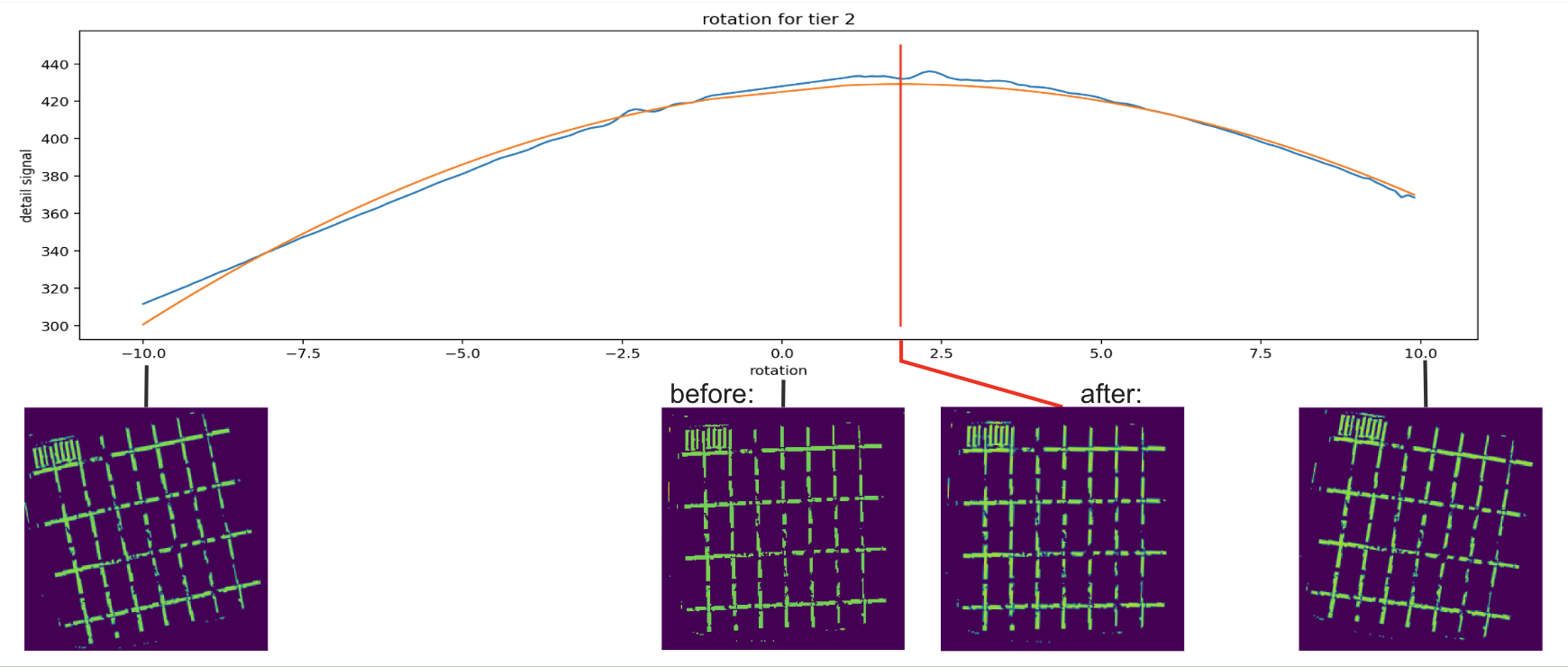}
    \caption{Automatic rotation. Top: vertical/horizontal to diagonal detail ratio ($J$) plotted in blue against the angle of image rotation $\theta$ in degrees. Angles were sampled every 0.1 degrees between -10 and 10. Due to noise about the peak, a quadratic fit (orange) was performed to find the peak rotation angle. Middle left: detected tier dividers image before rotation ($\theta = 0$). Far left: $\theta = -10$. Far right: $\theta = 10$. Middle right: output image corresponding to detected peak response from quadratic fit (about 1.9 degrees here).}
    \label{fig:auto_rot}
\end{figure}
\vspace{-1em}
{\bf Phase 5: Grid Segmentation} was done by summing the final rotated $B$ image along the $x$ and $y$ axes and identifying peaks, as shown in \figref{Figure}{fig:final_seg}. The CSV file conveys how many rows and columns there are, i.e. how many peaks we should seek for each tier. Extracting the box coordinates corresponding to nonempty cells in the CSV file and storing and rescaling the various crops and rotations along the way, we can then extract the fragments from the original scan. 

\begin{figure}[H]
    \centering
    \includegraphics[width=.9\textwidth, clip= true, trim = 7 5 2 2]{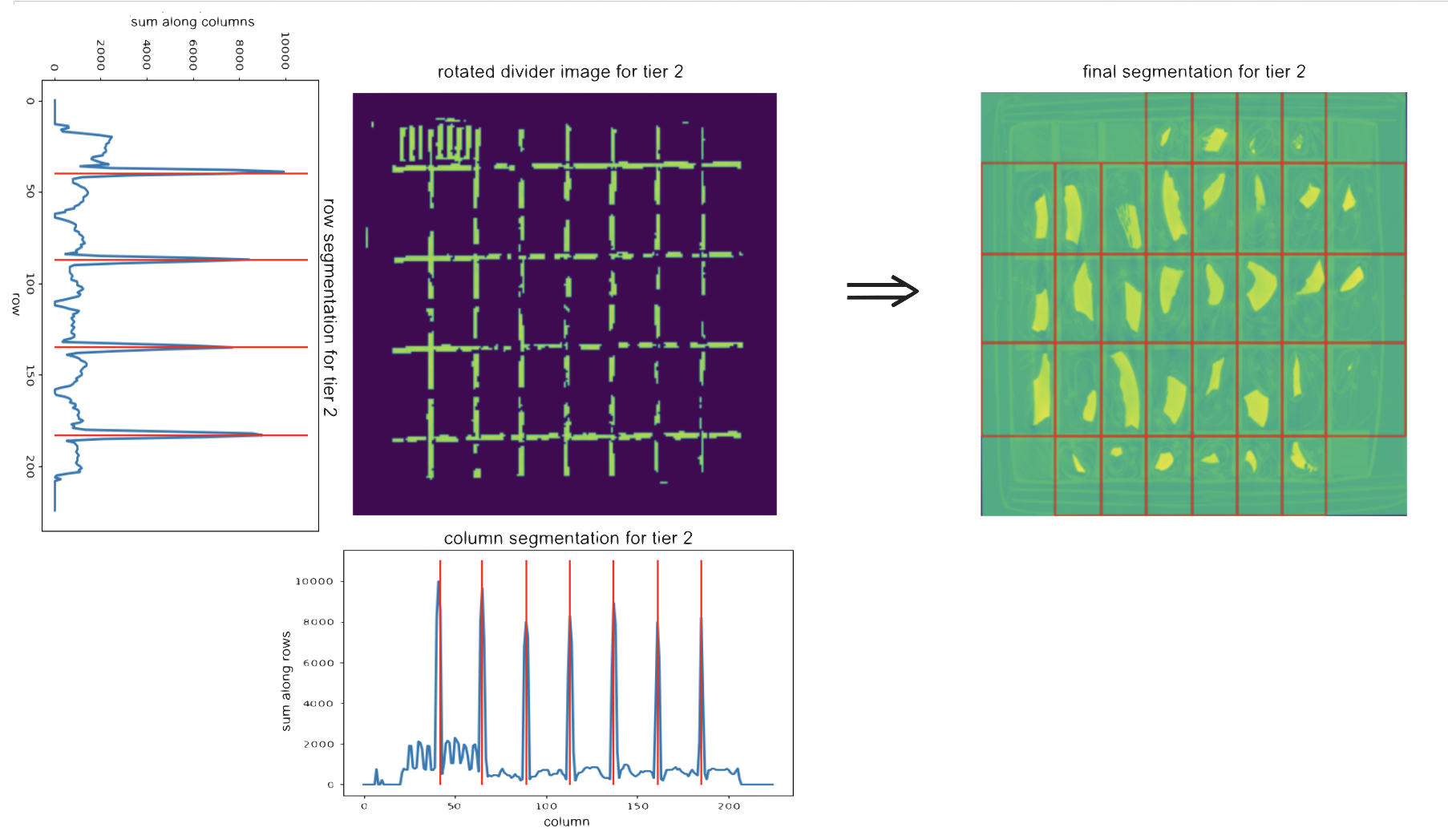}
    \caption{Final grid segmentation for a single tier.}
    \label{fig:final_seg}
\end{figure}

\subsection{Bounding box extraction, marching cubes, and cleaning}

The extracted bounding boxes were padding slightly so as to prevent possible truncating of any fragments. Due to memory constraints, this was done by looping over each object in the scan and extracting and storing the sub-array corresponding to the box coordinates. This is by far the most time-consuming part of the runtime, but is readily accelerated with parallelization. Surfacing was initially done using marching cubes with the second user-identified threshold (upper limit of cardboard range) as the input isolevel (the voxel value that objects will be at or above), and revised to a different isolevel if needed. Finally, the output meshes were cleaned. This was done by keeping the largest connected component without holes. Other approaches to cleaning may suit other tasks. 

\section{Results}
From 1,153 objects packed into just 10 scanning packages, we successfully scanned, processed, and surfaced 1,112 objects in entirety to a resolution of  41.8 - 56.0 $\ \mu$m. The resolution varied slightly from scan to scan. The disparity in the number of objects comes from the first (trial) scanning package being too large for the scanner’s field of view. This means some objects were not completely in the scan, as discussed more in \figref{Section}{sec:fov}{}; all other 9 scans fit within the field of view, and all objects in them surfaced successfully and in entirety. By packing the objects into a tiered 3D grid structure and creating CSV files that track the layout and identifier of objects in each tier, the scans were successfully segmented into individual cells and then surfaced into individual object PLY files with the correct object identifiers.

Scan packaging and documenting the layouts in a CSV file took about 30-45 minutes per scan. Each scan took 1-1.5 hours to complete, and preliminary post processing (manual) took 30 minutes per scan (done concurrently with scanning). The final segmentation and surfacing workflow was almost completely automated, requiring under 10 minutes of manual human intervention in just the early portion of the workflow (only steps 1 and 3; steps 4-5 are the bulk of the runtime). These algorithms took less than 4 hours to run (unparallelized), or 1-1.5 hours parallelized on a more advanced computer. Thus, to scan our entire sample of 1,112 fragments, it took 25-38 hours, of which only 11-14.5 hours required manual human interaction. This averages to 1.1-1.5 hours of manual human work per each scan (average of 115 objects in each). This is 35 seconds of human time per object and about 2 minutes total per object. \figref{Figure}{fig:bone_surf}{} showcases some of our resulting bone surfaces. 

\begin{figure}[H]
    \centering
    \includegraphics[width=\textwidth]{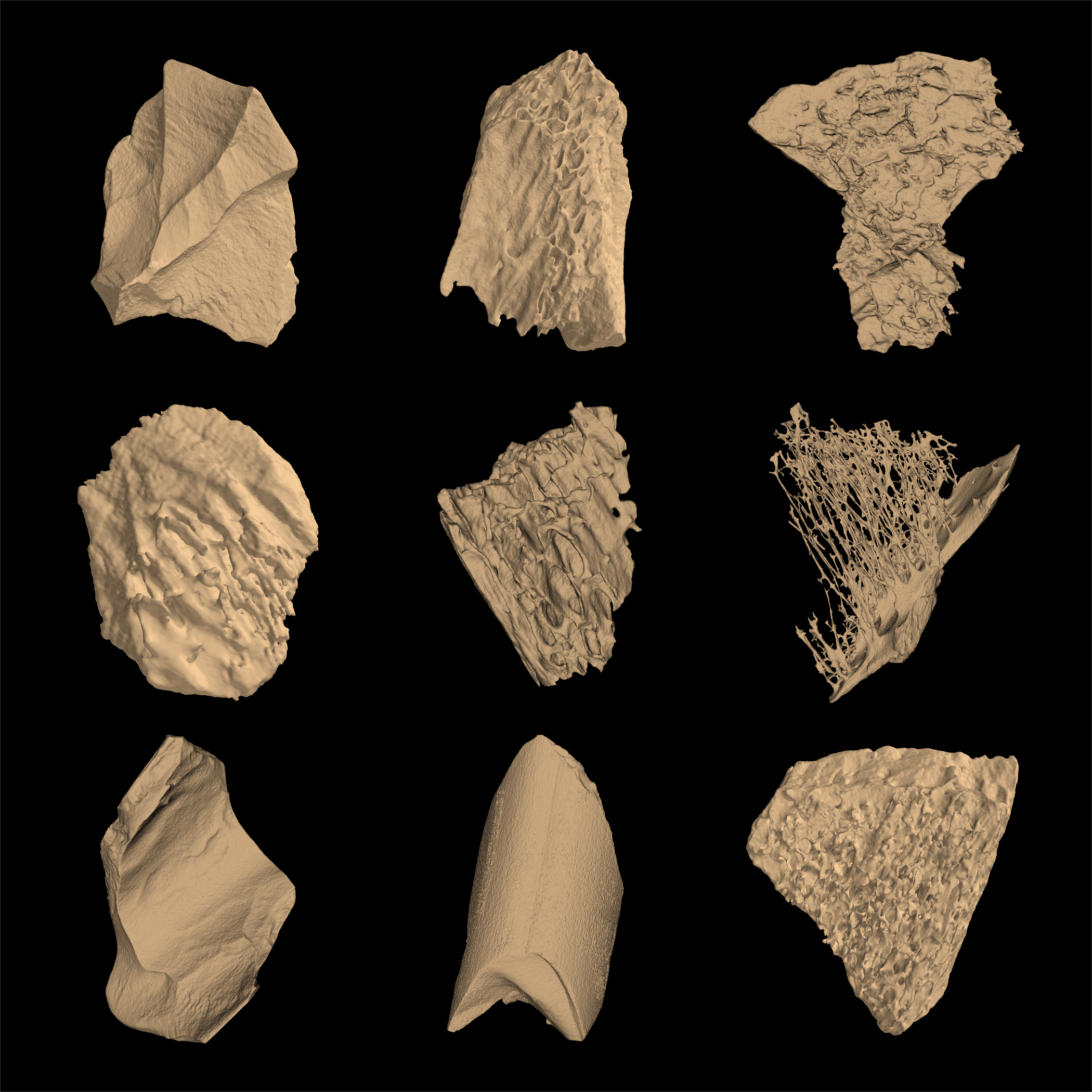}
    \caption{Final micro-CT surfaces of bone fragments, all with $41.8-56.0 \mu$m resolution.}
    \label{fig:bone_surf}
\end{figure}

\section{Discussion}

\subsection{Important Considerations When Scanning}

There is no single scanning method that will be optimal for all projects writ large. In order to choose which method to use when approaching a research project involving 3D surface modeling, there are several inter-related considerations to evaluate. The first level of consideration pertains to resolution and sample size. The sample size needs to be sufficiently large to address the research question and that the resolution of the 3D models must capture the detail necessary to address the research question. Once this is decided, attention can turn to the details of each stage of the workflow. Broadly, these stages are: preparing objects for scanning; scanning; processing data and surfacing scans; utilizing the 3D models for research; and curation of all data and metadata. The final major consideration is replication and reproducibility. The cost/benefit analysis can be thought of in terms of time (interactive and automated), money, human resources, required skill sets, location, materials, and scanning and computer hardware and software. 

\subsubsection{Resolution and Sample Size}
The use of high resolution scanning introduces challenges relating to data-structures (file format), computer memory and processors, storage, and processing time especially as it pertains to working with larger samples. For example, during processing, a typical scan from this project required 62 GB for just the TIFF images (scan slices), and all 10 scans' slices required 615 GB in total. With the associated scan metadata, final object surface PLY files, and intermediary NPZ subvolumes (compressed Numpy arrays) and other metadata produced by our workflow, this project consumed 2.7 TB of storage. While it is not essential to keep the NPZ subvolume files (2-6GB each) and just keep the output PLY surface files (8-64 MB each), we elected to for possible downstream tasks. All of this was stored on a 4TB drive for portability. 
Unlike previous work, our workflow never brings the entire scan into machine memory for processing. Paired with the portable drive, this allows our workflow to run on any standard consumer-grade computers with at least 16GB of memory (provided the extracted object sub-volumes are less than this - more on that momentarily). This allows those without access to intensive computing resources but still access to micro-CT scanning and suitable storage space to use our methods.


Subvolume extraction and surfacing via marching cubes are by far the most memory-intensive and time-consuming parts of our workflow, as these perform many operations on full-resolution subvolumes. After some experimentation to find the fastest procedure, we implemented a single slice reading loop where subvolume files are incrementally appended {\em in-disk} as opposed to {\em in-memory} using the recent npy\_append\_array module  \cite{npyappend}. Appending these subvolume arrays in-disk as opposed to in-memory eases memory constraints, allowing one to scan larger objects as well (e.g. 4 large objects as opposed to 150 small objects). Sub-image extraction and writing may be parallelized within the reading loop for faster execution. Unparallelized, the runtime for this took 3 hours for a representative scan, and one should expect even faster times if parallelizing the inner image-writing loop.


Surfacing via marching cubes was parallelized as well for faster execution, wherein each worker loads a subvolume into memory and surfaces it. Here we find the ultimate memory bottleneck of our workflow: the entire subvolume must fit comfortably into memory with room for performing the marching cubes algorithm. Most of our subvolumes were on the order of 5GB; for 16GB memory machines, parallelization is less tractable, and this can be time consuming. For machines with much larger memory and more CPUs, again parallelization contributes to a substantial speedup. For applications with relatively large objects at high resolutions (e.g. 16GB / object bounding box) and computers with limited memories (e.g. 16GB), ultimately a different approach to surfacing must be utilized, but we did not encounter this problem.

\subsubsection{Workflow Considerations} 

In order to prepare the physical objects for scanning, the first consideration is whether sample can be transported to the scanner or, if the scanner needs to be transported to the sample. In this case, we transported the objects to a large, fixed scanner. Portable micro-CT scanners are still extremely rare, though there are bench or desktop micro-CT and XRM scanners available through companies such as Bruker. Perhaps such a scanner could be purchased for permanent use at a field site but transporting the scanner back and forth might not be feasible. For example, one of Bruker's smaller XRM scanners, the SKYSCAN 1275, weighs 170 kg (340 lbs). 

However, the ability to transport the collection to the scanner can save the overall time investment and financial expenditures. Several hundred small objects can be scanned simultaneously using our method: scanning 150 objects individually with micro-CT would require more preliminary post-processing time overall and cost several thousand dollars (\$15,000 if for an hour each at our rates), but packing these together for a single scan with our method would cost only \$150. Further, the materials used for packaging the physical objects for transport and scanning were inexpensive and easy to acquire. 

Time is money too, and our method is extremely effective for use of human time, scanner time, and post-processing time. A package of 150 fragments only takes 1-1.5 hours to scan, whilst scanning 150 fragments individually takes well beyond this (not to mention load/unload time and added post-processing time). Our segmentation and surfacing protocol takes under 4 hours to run (unparallelized), and can leverage greater computational resources (i.e. more CPUs and RAM for parallelization) to drastically accelerate this to 1-1.5 hours or less. Only 1-1.5 hours of manual human work were required for each scan (30-45 min packing, 30 min preliminary post processing, and $<$ 10 min of input to the segmentation and surfacing workflow). Furthermore, as we developed this protocol, we experimented with the total number of fragments that could be included in each scan. In retrospect, a maximum of 158 could have been utilized, resulting in an overall sample of 1580 reducing the overall time per object to 90 seconds.

Though micro-CT scanners and scanning are expensive, following our protocol to batch many objects together in a single scan can deliver tremendous cost-savings per object. micro-CT scanners range in price from about \$250,000 to \$1,000,000 and are often purchased for use by many organizational divisions and projects so the costs to individual projects are reduced. Our scans were acquired at the University of Minnesota's XRCT lab, which charged \$150 per scan. The total cost of our 10 scans after packaging materials was \$1,549.99. While this is a large sum of money, with 1,112 objects successfully scanned, the price per object is only \$1.18! If filling to the maximum of 158 objects per scan, this is only \$0.98 /object. Furthermore, once the 3D models are acquired their use life can be extended at no extra cost through use in future projects and via data sharing. 

\subsubsection{Curation, Replication, and Reproducibility}

The capacity to create voluminous amounts of data is only the first step. Handling large amounts of data after-the-fact can be a complex process. This is especially true within archaeology, where researchers must synchronize and work between collections in both their physical form and digital form along with the associated metadata that provides the all-important context necessary in our field. Furthermore, excavation and research at archaeological sites are generally ongoing meaning that collections (physical and virtual) oftentimes continue to expand over time. The sheer amount of data and metadata can be confusing and overwhelming, especially for those who are accessing these collections with minimal prior experience with either the collections or the methods utilized to create them (e.g. students, early-career researchers, independent research groups, and educators). Such expansion in the sheer amount of data coupled with the goal of optimizing access to broad audiences amplifies the curatorial and infrastructural needs for building organizational frameworks that support the usability of shared data. 

Our new Micro-CT protocol offers a simple, (comparatively) affordable, and user-friendly means to package and track objects through the workflow, as well as organize surface data afterward. Provided curatorial standards and object identifiers that can accommodate a growing collection, our packaging and scanning protocols are readily deployable and intuitive. The output surfaces are named with the object identifier and isolevel, which allows easy organization after scanning as well as the user to experiment with multiple isolevels. Each step of our post-processing saves auxiliary meta-data files that track coordinates, rotations, and important thresholds, so the workflow is entirely replicable (and may reuse such values). All our code will be published on the \href{https://github.com/jwcalder/AMAAZETools}{AMAAZETools Github} as well as linked to from  \href{https://amaaze.umn.edu/}{our website}. Models from this project will be available on request and we aim to begin putting them in an open-access repository by the end of Dec 2024.

\subsection{Lessons We Learned Along the Way}

The scientific method is inextricable from trial and error. Here we outline some pitfalls we encountered, our workarounds, and means of eliminating such issues in future uses of our protocol.

\subsubsection{Voxel Value Distibributions and Optimal Isolevels for Surfacing} 
\label{sec:vox}
After scanning and preliminary post-processing, we found that the distribution of micro-CT voxel values varied drastically from scan to scan. Unlike CT scanners, which are calibrated to output Hounsfield units (and which \cite{yezzi2022batch} used to develop the completely automated CT batch scanning protocol), micro-CT scanners are not calibrated to be in Hounsfield units - the output voxel values are extremely sensitive to a number of parameters for each scan, including KV MicroAmp, table position, detector position, beam pardoning, and more. This means that the output voxel values can vary wildly between scans, e.g. bones could be 14000-26000 in one scan and 32000-42000 in another, although the two distributions share the same characteristics outlined in \figref{Section}{sec:thresh}{}. As such, it is inordinately difficult to calibrate the output voxel values without some controls in each scan of a known density, e.g. a vial of water. We did not do this, and rather than try to match voxel distributions between scans, we opted for simple manual inspection. However, if placing controls in each scan, this could easily be automated. 

Therein, the optimal isolevel for surfacing varies between scan and often requires some trial and error to select. Were the scans calibrated and standardized, it would be easy to specify the isolevel corresponding to the objects' physical density/ies. Surfacing is relatively fast to rerun for each scan, so several values may be inspected to identify the user's preference. The ideal isolevel is very important, especially when one's objects are thin. Thus, in situations where the objects may vary greatly in thickness, it is useful to package together items that have a similar thickness.

\subsubsection{Capturing the Entire Specimen}  
\label{sec:fov}

While our first trial scanning package fit on the Micro-CT {\em scanning bed}, it did not fit in the scanner's {\em field of view}. As such, 39 objects from scan 1 failed to yield complete surfaces. Our subsequent packages were smaller to fit within the field of view. When first implementing a high-throughput micro-CT scanning workflow, the scanner's field of view and corresponding resolution should be ascertained prior to starting scanning so an appropriate package size may be used.

\subsubsection{Cell Alignment within Package Tiers} 
Few objects in this study filled their entire cells. As space is money in this setting, an organizing system with an irregular arrangement of rectangles (e.g. at right in \figref{Figure}{fig:grid-layout}{}) may seem more efficient. While this can be scanned, this will not work with our CSV scan-layout organizational system nor grid detection algorithm. Our protocol needs objects to be arranged in an $N\times M$ grid. The width of each row and column may vary, but the grid lines must go from end to end in the box. An acceptable example of this is shown at left in \figref{Figure}{fig:grid-layout}{}. Therein, it is prudent to organize like-sized objects together before packing so one may get the maximal number in a given scan.

\begin{figure}[H]
    \centering
    \subfloat{\includegraphics{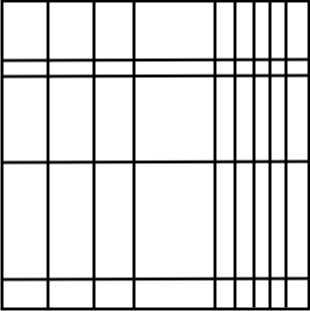}}
    \hspace{40pt}
    \subfloat{\includegraphics{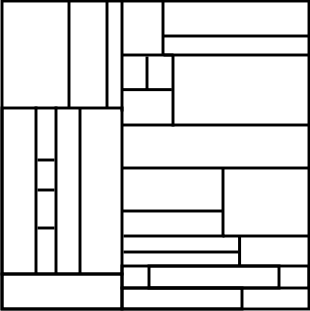}}
    \caption{Examples of acceptable and unacceptable tier arrangements. Left: an acceptable grid arrangement for our protocol. Objects are arranged in an $N\times M$ grid; the width of the rows and columns may vary within a tier, and the number of rows and columns may vary between tiers in the same scan. Right: while this may be acceptable as a Mondrian inspired design, it is an unacceptable grid arrangement for our protocol. Grid lines must go side to side and yield an $N\times M$ arrangement to store and recover the object identities.}
    \label{fig:grid-layout}
\end{figure}

Note that the number of rows ($N$) and columns ($M$) need not be fixed for all tiers in a given scan, and indeed the tier height may vary as well, allowing for more flexible and efficient scanning. For all but one of our scans (CT 1), we used the same grid arrangement for all tiers. Crucially, if electing for variable $N$ and $M$ within the same scan, the number of rows and columns should also be stored for each tier in the scan-layout CSV file so as to enable our fully automated grid recognition code. This would require but a minor modification to our code and previously presented scan layout CSV arrangements.

\subsubsection{Future Research and Applications}
Future research largely spans ways to make the algorithm completely automated and more efficient. As discussed in \figref{Section}{sec:vox}{}, known density controls (e.g. vials of air or water) could be placed in the scan to calibrate the output voxel distribution. This could make bounding box identification and scan orientation completely automatic and give known isolevels corresponding to the objects of interests' density. Further, an interlocking 3D tier grid would remove the need for automatic tier rotation, and remove some tier-wise operations as well. More advanced or automatic methods for isolevel selection merit exploration as well.

As to future applications, our protocol rapidly increases capacities for scanning large scale archaeological and paleontological collections, enables broader access to materials by independent research teams, promotes digital curation and expanding public outreach in museum settings, invites increasingly robust applications using methods such as machine learning and deep learning, and opens pathways for advancing digital data extraction within archaeology. Furthermore, our protocol and workflows may be immediately adopted by scanning institutes, centers, and companies to batch several customers’ orders together and achieve lower scanning cost, e.g. closer to \$0.98/object opposed to \$150 for scanning an object individually. This would mark a tremendous development for 3D modeling and democratize high-resolution scanning. Our protocol also works at a larger scale as a means for en-masse CT scanning, where a human-sized box or several boxes can be used to pack hundreds to thousands of objects. Further, as our workflow applies to any objects with a density discernible from the dividers and cardboard, a plethora of objects beyond bones may be scanned, including lithics, ceramics, metals, glass, circuits, and more. Therein, many other fields may immediately apply our methods to implement their own mass-scale micro-CT or CT scanning studies.

\section{Conclusion}
Our new en masse micro-CT scanning protocol offers a novel and efficient means for obtaining 3D models for hundreds, if not thousands, of small objects from a single scan. From only 10 micro-CT scans, we successfully obtained high resolution 3D models for 1,112 objects at a cost of just \$1.40 USD each, or \$0.98 USD if we had packed them to the fullest. Building off previous semi-automated approaches to batched medical CT scanning \citep{yezzi2022batch}, our work here offers hassle free packaging, mostly automated and memory-efficient post-processing with minimal human intervention, and tectonic cost and time savings all around. Notably our post-processing workflow runs on any computer with at least 16GB of RAM, taking less than 4 hours per scan on 1 CPU. This enables those with sufficient storage but limited computational resources to still use our methods, therein making high-resolution scanning more accessible. With more CPUs and RAM, the memory-intensive parts of our workflow are easily parallelizable to deliver runtimes on the order of 1-2 hours per scan, therein increasing the accessibility to high-resolution scanning. While human intervention in post-processing takes less than 10 minutes per scan, we outlined ways to make post-processing fully automated. Likewise, automated scan packaging and labelling could be done for industrial applications. 

As our method works for any objects with densities discernible from the cardboard and plastic dividers, it applies to innumerable types objects (provided the objects all fit within the chosen grid system and scanner’s field of view). Therein, this bears immediate applications in numerous fields, including paleontology, geology, engineering, materials science, and manufacturing. Our \href{https://github.com/oneil571/AMAAZE-MCT-Processing}{open-source code repository} and convenient packaging protocol allows our method to be rapidly adopted for such applications. 

Beyond use for en masse micro-CT scanning, our protocol immediately works as a method for en masse medical CT scanning. Several boxes or a single human-sized box packed in the same way we described may be scanned to surface tens of thousands of objects simultaneously. In addition to use by other mass-scale micro-CT or CT scanning studies, our method may also be adopted by micro-CT scanning centers to batch several customers' orders together and scan them simultaneously. This would offer tremendous consumer cost savings, greater efficiency, reduced power usage, and greater accessibility to high-resolution models for all. 

As our world is increasingly digital and studies demand increasingly higher resolution data, increased efficiency and accessibility of micro-CT scanning are paramount. Here, we introduced a new, scalable, open-source method for mass scanning that enhances researchers' ability to leverage high-resolution data. This approach not only benefits anthropology but across disciplines in a way that democratizes access to advanced research using micro-CT and offers a significant leap in efficiency within archaeology, a discipline that uses extensive material collections. 

\section{Acknowledgments} RCWO was supported by NSF GRFP-2237827 and NSF-DMS:1944925, KYW was supported by NSF SPRF-2204135, and JC was supported by NSF-DMS:1944925 and NSF-CCF:2212318. We are grateful to insightful conversations with John Brigham of the University of Minnesota Visible Heart Laboratories. The bones were acquired from Scott Salonek with the Elk Marketing Council and Christine Kvapil with Crescent Quality Meats. Bones were broken by hyenas at the Milwaukee County Zoo and Irvine Park Zoo in Chippewa Falls, Wisconsin or by various math and anthropology student volunteers who broke bones using stone tools. Abby Brown and the Anatomy Laboratory in the University of Minnesota’s College of Veterinary Medicine provided the facility to clean bones fragments. The bone fragments are curated by Matt Edling and the University of Minnesota’s Evolutionary Anthropology Labs. The fragments were scanned at the Department of Earth and Environmental Sciences XRCT lab with the guidance of Francesca Socki. 

\bibliography{references}
\bibliographystyle{apacite}

\newpage
\section*{Supplemental Information}

\section{Sample In Detail}
This table shows the full list of sample identifiers, the dimensions of the fragments (in mm), and their volumes ($(mm)^3$)to give a detailed view of the size of the objects and to inform future reproducibility. Though some may be interested in knowing species and skeletal part of these fragments, that is not the focus of this study; such information will be made available when the models are shared. 

\newgeometry{left=.5cm,right=.5cm,bottom=3cm}

\begin{center}    
\begin{small}
\begin{longtable}{|l|l|l|l|l|l|l|l|l|l|} 
\caption{List of Specimens} \label{tab:specimens}\\
\hline
Specimen & Length & Width & Depth & Volume & Specimen & Length & Width & Depth & Volume  \\
\hline
\endfirsthead

\multicolumn{10}{l}
{{\bfseries \tablename\ \thetable{} -- continued from previous page}} \\
\hline
Specimen & Length & Width & Depth & Volume & Specimen & Length & Width & Depth & Volume \\
\hline
\endhead

\hline 
\endfoot

\begin{filecontents*}{mesh_stats.csv}
specimen,length,width,height,volume,specimen,length,width,height,volume
MN19061107,16.261,3.206,1.764,22.982,MN18060315,18.781,14.931,5.054,315.036
MN19061005,33.583,8.507,7.506,495.958,MN19065313,34.161,12.653,4.719,386.377
MN19081508,12.586,3.585,2.117,16.799,MN19063910,27.192,9.441,5.253,218.029
MN18020230,35.495,7.83,3.45,222.423,MN19064914,22.754,10.604,3.436,159.139
MN18021815,12.968,9.771,3.441,125.356,MN19064209,38.388,9.803,4.375,718.229
MN18020117,37.568,21.229,7.235,1754.648,MN18061108,25.128,14.599,9.26,822.62
MN19061112,15.391,9.401,2.273,102.126,MN19063608,16.928,12.261,4.781,296.197
MN19061007,31.533,11.992,4.355,401.301,MN19064913,17.867,9.204,4.087,181.218
MN18020233,31.293,9.197,5.045,352.261,MN19063806,46.475,19.294,9.001,1047.218
MN15011012,11.912,5.265,1.595,26.277,MN19063610,17.631,8.487,3.867,104.51
MN14AA0425,21.113,6.731,4.791,126.232,MN19064507,27.615,11.619,5.329,455.431
MN18020247,31.925,7.722,4.553,301.297,MN19064810,17.014,10.672,3.887,194.981
MN18020611,30.58,12.524,5.357,541.293,MN18060726,21.558,13.357,6.626,347.994
MN15011007,10.472,9.245,4.331,97.593,MN19063908,26.311,11.279,4.261,372.021
MN19081414,20.341,10.173,5.974,356.665,MN18061015,26.629,10.249,5.644,472.464
MN19061808,20.513,8.162,5.505,199.699,MN19063607,20.915,13.491,5.726,400.51
MN18020225,45.039,11.82,6.07,1131.994,MN18061106,33.067,18.887,8.323,931.653
MN18020237,27.659,5.505,4.108,202.07,MN19065308,43.655,18.957,7.245,1302.109
MN19061608,26.06,10.374,4.891,304.988,MN18060722,27.583,17.446,3.913,405.766
MN19062004,27.265,10.247,5.066,460.189,MN18060711,42.527,18.428,8.353,1620.726
MN14AA0522,35.375,9.728,7.498,939.048,MN18061618,17.16,10.385,3.727,129.373
MN18021714,11.994,6.431,2.178,39.066,MN18060806,28.447,8.571,4.962,397.871
MN19061215,16.503,6.571,4.097,100.554,MN18060713,32.117,21.727,9.852,1103.509
MN19062111,12.523,5.373,2.192,37.342,MN18061311,29.356,19.383,5.181,563.452
MN15AA2409,20.623,19.309,6.395,350.8,MN19064004,29.711,8.811,4.63,278.006
MN15010407,28.813,7.857,5.256,277.444,MN19064302,49.661,20.91,9.327,1584.615
MN19062015,37.126,11.965,6.494,790.476,MN19063604,42.484,19.853,9.821,1502.113
MN15010817,8.697,6.609,2.971,43.946,MN19064212,33.882,16.318,5.152,551.985
MN14AA0429,19.081,6.82,6.198,176.067,MN19065028,20.897,6.026,2.533,46.768
MN19062104,33.226,14.408,9.4,1087.99,MN18060809,31.318,11.611,6.572,531.38
MN19081421,33.272,13.742,9.071,743.457,MN18061508,33.465,13.391,6.273,924.858
MN19081506,13.764,7.072,3.521,88.798,MN18060727,24.529,10.566,5.877,361.817
MN18020112,41.452,14.961,5.517,980.611,MN19065411,19.432,8.117,3.079,155.959
MN15AA2408,23.298,17.271,5.456,439.843,MN18030515,26.036,13.225,5.989,563.475
MN19060908,1.58,1.26,0.408,0.304,MN18030514,31.014,9.873,6.477,502.428
MN14AA0322,24.941,9.416,7.014,484.558,MN18080116,21.244,10.034,3.801,235.115
MN19060804,48.347,16.464,6.766,1315.805,MN19081218,20.304,12.126,6.212,548.744
MN19060903,33.071,15.017,8.543,1424.733,MN18080112,48.339,18.5,7.149,1672.388
MN18020109,37.319,20.432,5.864,1236.54,MN19081128,12.887,11.958,5.222,200.322
MN18020833,38.013,12.582,7.609,724.074,MN18030519,29.37,11.827,4.025,287.651
MN19060103,20.133,8.972,5.022,217.291,MN19065105,31.817,20.807,8.045,1129.848
MN14AA0326,26.426,7.368,5.261,210.023,MN19050626,24.604,8.022,7.651,436.281
MN19062304,11.741,8.965,4.555,82.762,MN18030516,27.509,22.368,8.826,355.784
MN19061108,0.994,0.538,0.318,0.066,MN18080117,6.024,2.308,0.483,0.887
MN15010815,14.146,5.009,3.035,56.532,MN18030713,20.973,14.157,5.682,299.842
MN18020246,21.399,9.53,2.894,113.912,MN19050622,2.635,0.765,0.69,0.161
MN18020116,14.37,8.006,2.461,92.789,MN18080113,38.07,16.02,6.206,907.944
MN19081313,24.306,6.371,3.812,147.982,MN18030309,37.013,16.992,11.046,847.642
MN15AA2407,25.207,12.257,4.218,250.189,MN18030814,32.886,13.347,5.806,375.956
MN18020721,18.144,7.632,5.103,221.55,MN18030612,24.623,10.09,7.455,230.288
MN14AA0526,25.115,11.495,6.521,508.459,MN19081207,31.902,13.486,6.391,801.909
MN18020244,28.744,8.158,5.1,320.886,MN18030815,20.026,12.946,5.671,388.576
MN19060911,29.725,10.422,5.927,465.167,MN19065410,30.647,10.88,4.817,626.048
MN13AA0710,67.835,15.016,7.685,2819.52,MN18030813,19.709,16.899,9.121,641.864
MN13AA0810,19.859,11.215,2.818,162.196,MN18030310,38.344,12.633,7.892,813.195
MN18020232,49.616,17.124,9.608,1494.086,MN18030715,24.962,11.251,8.023,475.287
MN19060907,33.35,16.616,5.704,1062.67,MN19050628,2.635,0.765,0.69,0.161
MN19061209,7.868,1.853,1.151,0.631,MN18030809,37.311,15.991,9.243,1412.479
MN13AA0718,25.656,6.815,1.795,46.307,MN18030211,13.747,13.255,6.128,310.141
MN14AA0524,13.991,10.156,3.249,94.455,MN18030210,29.915,12.896,7.54,909.533
MN19061211,18.879,8.469,4.554,270.151,MN19065205,14.723,10.437,2.9,123.811
MN18020248,21.739,13.321,6.403,477.318,MN18030609,25.395,21.428,8.426,1298.089
MN19061217,21.953,7.22,3.374,157.025,MN19081217,21.034,10.746,4.904,351.232
MN19062405,49.035,14.023,7.449,1401.173,MN19050632,17.915,7.702,3.535,131.967
MN19060912,39.104,6.767,4.822,423.019,MN19064605,43.112,19.624,9.906,1505.847
MN19061409,17.864,8.862,3.753,127.95,MN19080704,18.351,9.891,4.02,229.443
MN19062012,10.563,6.646,3.717,89.239,MN18030712,18.339,14.444,7.495,600.775
MN19061113,33.306,10.278,3.507,286.74,MN18030816,17.896,12.597,5.741,190.111
MN19061811,17.838,7.695,2.4,60.994,MN19065108,21.401,9.873,4.293,258.51
MN18020827,29.492,12.132,7.723,824.327,MN19064215,18.701,6.956,3.594,131.828
MN13AA0711,42.549,20.22,10.019,2024.682,MN19064412,18.924,5.894,2.629,57.488
MN19081318,0.533,0.463,0.186,0.011,MN19063810,21.791,7.623,4.531,214.416
MN18021713,26.531,12.796,5.296,374.131,MN19050631,15.071,9.16,6.094,239.543
MN19061216,10.574,10.422,5.066,172.588,MN19081210,37.812,10.217,5.805,628.084
MN18020234,23.967,6.28,3.16,116.21,MN18030714,24.153,16.91,7.884,440.124
MN13AA0716,40.763,14.013,7.058,938.559,MN18030719,15.305,12.709,5.214,291.178
MN18020822,25.91,9.683,3.94,299.494,MN18030110,29.208,7.474,5.444,433.322
MN15AA2504,33.721,16.899,7.357,903.235,MN19064410,14.169,8.034,6.355,137.675
MN19062403,44.041,13.276,5.833,916.31,MN18030414,30.18,20.043,8.21,1319.292
MN19062109,57.792,8.323,5.482,666.036,MN18080114,44.84,12.907,7.652,873.739
MN19062009,16.834,11.637,4.506,175.982,MN19065409,36.32,12.484,5.78,387.324
MN18020830,45.834,9.985,5.239,555.174,MN18030906,39.714,23.4,11.698,2881.996
MN15010611,23.917,6.994,2.098,88.231,MN19081110,36.513,10.998,4.266,435.64
MN19060505,38.065,10.191,8.392,1279.269,MN18030418,33.635,10.773,7.915,440.166
MN14AA0323,13.601,9.281,2.226,70.052,MN18080110,44.343,19.192,7.68,1557.382
MN19060803,20.329,13.325,8.709,140.966,MN19081222,23.592,12.513,7.187,315.008
MN15AA2410,16.612,6.787,2.18,71.873,MN18030711,25.433,13.076,7.394,556.284
MN18020720,19.078,7.49,4.219,257.64,MN18030717,17.477,14.293,4.607,381.812
MN19081413,44.035,14.263,6.404,1352.274,MN19050627,20.105,5.776,3.032,111.248
MN19060705,36.433,15.667,6.667,973.011,MN19081204,46.111,20.523,9.069,1425.81
MN19062406,30.508,11.922,4.862,630.763,MN18030108,24.16,13.81,5.618,477.135
MN19060205,62.579,20.444,11.671,2646.75,MN19081211,39.681,15.795,8.932,540.726
MN13AA0709,34.649,23.773,8.913,1434.955,MN19081212,37.684,8.022,4.969,393.092
MN19081412,31.164,16.931,10.824,697.314,MN18030311,35.402,9.666,6.137,519.775
MN19062014,26.531,6.502,3.515,181.072,MN19081114,28.787,14.97,5.565,450.886
MN19062006,26.475,9.508,4.049,176.468,MN19080703,20.809,15.658,7.385,479.127
MN19061412,36.71,7.762,3.761,214.094,MN18030716,23.688,12.939,5.962,380.759
MN19061610,18.182,10.178,6.042,368.464,MN18030610,24.726,14.579,9.256,538.126
MN19061414,16.526,5.497,4.347,92.235,MN18030720,21.613,10.458,5.584,324.789
MN19061114,26.947,9.186,4.706,256.765,MN19064216,18.85,4.841,3.516,105.576
MN19061605,27.288,9.954,3.396,171.565,MN18030109,27.743,8.788,5.351,233.305
MN18020609,10.438,9.627,4.647,102.883,MN19081221,22.422,11.625,8.356,512.424
MN19062005,68.259,16.825,9.604,2068.275,MN18080118,17.852,8.846,4.866,184.223
MN19081317,1.963,0.671,0.307,0.14,MN19081216,19.536,9.932,8.804,376.565
MN19061810,14.351,6.22,3.224,67.228,MN19050634,18.989,5.787,2.99,54.152
MN19062011,15.359,9.412,3.327,74.362,MN18030611,23.373,12.007,4.294,328.112
MN19061214,16.189,3.552,1.314,15.133,MN19065018,14.09,11.286,5.945,192.577
MN19062105,47.354,15.028,6.3,1368.341,MN18080111,46.622,20.319,10.816,2089.552
MN19061410,24.815,8.793,3.513,158.715,MN18030518,27.496,10.739,5.628,425.489
MN19081310,25.414,10.359,4.635,177.646,MN18030718,19.006,11.716,5.545,444.963
MN19061106,48.821,9.064,8.477,767.485,MN18030723,13.574,9.861,2.921,104.945
MN18020221,55.395,16.627,5.537,1649.318,MN19081209,42.328,12.259,8.203,1606.17
MN15010816,19.777,6.68,6.392,151.09,MN18030721,19.768,11.499,3.595,165.214
MN14AA0435,20.528,8.107,2.327,119.628,MN18030708,43.507,20.221,9.622,2580.686
MN19081418,38.946,12.978,7.337,932.435,MN18030413,39.538,19.192,10.504,2045.997
MN19081415,30.561,15.882,5.284,577.526,MN19081205,49.143,13.53,7.479,1644.434
MN18020250,25.509,9.498,6.081,441.916,MN19050635,14.611,6.898,5.481,200.984
MN13AA0812,26.627,9.92,3.681,279.198,MN19064606,34.545,13.037,7.741,675.152
MN19081505,21.932,11.632,5.583,209.385,MN190812119,23.241,10.722,4.8,411.485
MN18020820,30.348,12.515,7.028,724.466,MN18080115,24.878,10.813,4.352,353.45
MN19061210,46.461,11.188,5.239,799.867,MN18030811,26.831,18.29,4.881,470.813
MN18020920,23.08,3.762,1.42,23.057,MN19081223,24.88,10.203,3.801,358.868
MN15011005,18.27,8.668,2.668,101.966,MN18030613,22.586,8.202,6.931,370.504
MN18021812,18.23,6.296,2.445,78.772,MN19081213,35.905,9.593,6.421,513.989
MN18020908,18.874,13.163,5.02,387.465,MN19081215,24.436,13.652,6.746,534.806
MN19081315,12.177,8.285,3.03,80.313,MN18030923,15.008,8.527,3.138,124.123
MN18020831,21.455,10.238,6.433,277.686,MN18030722,15.14,9.078,5.084,151.642
MN19060808,23.09,8.079,3.73,176.829,MN19081220,23.299,15.005,1.998,166.401
MN18020110,36.618,22.173,8.765,2008.994,MN18030608,33.553,13.803,7.509,1345.535
MN15AA2507,13.329,8.722,3.594,84.123,MN19065107,32.929,13.85,8.486,601.666
MN13AA0723,35.307,15.173,5.607,542.907,MN18030308,52.183,18.095,9.998,1480.021
MN18021306,0.183,0.162,0.107,0.001,MN19063809,19.758,8.422,3.002,136.917
MN18020242,35.553,12.052,7.709,585.236,MN19064411,17.564,6.473,3.312,88.6
MN19060504,23.046,12.572,7.289,784.541,MN18030709,32.578,18.4,8.537,1208.685
MN19061807,40.055,9.639,5.222,303.131,MN18090211,47.604,13.637,5.94,1140.891
MN18020238,11.761,5.417,1.412,24.463,MN18080711,36.904,10.531,4.274,526.459
MN18020231,21.66,10.253,3.88,290.393,MN18080512,22.507,18.519,5.343,591.776
MN19061105,22.472,6.98,2.271,58.745,MN19081037,27.028,6.663,5.182,307.176
MN13AA0726,39.073,25.488,9.231,1781.94,MN19080938,31.287,18.792,9.773,1659.02
MN19062110,62.387,11.857,7.809,1808.065,MN19050612,38.54,17.61,6.583,1977.726
MN19061809,12.469,8.208,3.998,139.255,MN19050244,37.837,14.521,9.547,1937.579
MN18020245,20.137,9.254,4.267,168.343,MN18080617,21.173,12.891,4.611,361.898
MN18020227,41.478,14.102,5.755,939.9,MN19050259,19.165,8.558,4.963,211.095
MN18020220,63.339,15.059,6.896,1738.604,MN19080939,37.464,11.43,4.256,721.728
MN13AA0917,18.008,9.373,4.425,118.321,MN19081027,29.99,8.028,5.492,407.894
MN13AA0724,38.795,12.012,7.564,1226.391,MN18080416,45.097,20.43,7.681,1429.7
MN14AA0324,42.238,12.773,6.364,721.924,MN18080417,54.221,16.815,7.964,1931.941
MN19061207,33.825,14.275,3.726,335.618,MN18080513,26.19,15.136,6.304,596.272
MN19081316,14.579,3.416,1.14,13.583,MN18080621,17.367,8.321,4.331,154.931
MN18020829,39.51,10.712,3.652,333.821,MN19050254,24.229,14.543,6.623,532.292
MN19081410,28.315,11.805,4.523,324.706,MN18090807,49.431,12.12,6.217,958.14
MN18021814,10.892,5.928,2.656,40.393,MN11AA0504,34.889,13.485,8.504,779.056
MN14AA0521,45.672,10.476,5.329,671.033,MN19050630,15.635,12.828,5.276,189.964
MN19061413,14.142,11.028,4.346,139.584,MN19050245,35.107,14.865,7.167,1059.496
MN13AA1025,19.72,10.227,3.733,139.029,MN19050623,27.325,14.594,3.962,225.882
MN18020118,11.53,10.426,4.59,189.233,MN18090216,25.937,12.378,6.918,736.193
MN15010814,27.851,7.245,4.748,235.933,MN18080423,34.741,12.389,6.439,786.324
MN19062010,45.387,6.079,3.358,203.455,MN18080432,19.792,10.008,6.332,342.151
MN19081319,8.44,1.895,1.238,4.412,MN18080428,25.234,10.272,4.536,288.947
MN19061609,23.042,12.994,8.034,493.595,MN19050258,36.306,7.585,4.907,607.243
MN18020911,12.224,8.756,3.931,125.804,MN19050611,41.131,19.054,7.883,1641.557
MN18020723,12.2,1.07,0.81,1.314,MN19050610,42.315,18.504,5.193,1273.059
MN19060910,25.186,10.538,7.965,582.433,MN19050811,34.848,12.091,6.149,728.147
Mn19062103,45.089,13.989,8.305,875.368,MN18090309,50.855,14.885,7.463,1761.297
MN19060909,54.666,15.469,11.994,1426.552,MN19050545,35.871,13.229,6.421,991.649
MN19081503,36.543,10.018,4.412,333.272,MN18090214,41.401,15.351,6.836,1463.304
MN19061408,35.844,8.392,6.307,778.512,MN18080616,25.898,12.378,5.844,464.639
Mn18021305,8.414,4.196,1.004,7.166,MN19081020,34.748,12.694,6.298,1089.365
MN13AA0712,35.658,14.8,6.938,976.959,MN19050621,30.306,9.428,5.413,403.776
MN18020229,41.548,19.802,9.934,2242.375,MN19080936,37.382,12.082,6.57,937.655
MN19062112,39.984,8.63,2.903,203.945,MN19050552,30.904,11.685,7.84,759.291
MN19081419,24.671,10.602,5.043,450.023,MN19050813,32.47,10.917,7.195,456.06
MN18020228,9.255,5.84,1.614,4.988,MN18080510,51.77,16.309,7.307,1280.794
MN19062013,24.542,8.146,4.4,223.176,MN19050636,12.371,9.32,4.396,117.798
MN13AA0720,38.233,16.918,8.864,1409.231,MN19050124,36.573,8.709,7.182,868.322
MN19060905,30.848,15.098,6.302,671.644,MN11AA0505,38.803,10.362,7.84,780.19
MN15AA2411,26.126,12.055,4.42,193.424,MN19081021,33.217,14.731,7.207,1106.181
MN19060404,36.941,10.173,6.111,617.047,MN19050608,40.601,19.887,8.254,1806.61
MN19062107,48.044,21.177,10.765,2492.061,MN18080435,19.908,5.809,3.43,89.452
MN13AA0813,44.567,10.944,6.183,945.027,MN19080940,35.491,13.645,5.099,652.735
MN15010509,26.752,7.017,2.943,138.203,MN19050617,25.69,14.515,5.443,585.006
MN18020722,37.894,7.15,5.039,395.341,MN19050540,36.779,17.163,6.987,1340.472
MN19061115,45.712,8.982,4.179,405.708,MN18080434,16.926,8.866,3.814,137.547
MN19061213,3.23,1.926,0.62,0.593,MN19050551,31.291,13.7,10.294,970.181
MN19081420,27.27,11.951,5.993,465.184,MN11AA0807,32.089,10.726,4.495,266.925
MN19062008,17.856,9.969,4.69,261.487,MN19081023,32.768,11.597,6.286,730.524
MN14AA0327,60.322,12.941,5.535,665.252,MN19050262,22.819,9.587,3.425,146.15
MN19062407,28.798,9.886,2.865,245.424,MN18080419,50.061,17.439,9.442,2069.184
MN18020719,20.456,7.886,4.966,324.411,MN19081026,34.405,11.377,7.247,597.484
MN19061006,40.253,10.201,3.658,507.145,MN19050607,25.283,5.729,2.213,18.5
MN13AA0715,23.081,6.415,2.23,44.973,MN18080424,43.931,12.058,5.383,1024.238
MN19081320,17.705,7.117,5.242,141.754,MN18080511,34.948,19.203,7.247,1023.059
MN19060706,13.367,9.793,4.637,144.179,MN19080845,29.588,13.378,6.145,886.963
MN13AA0717,25.684,10.162,4.635,246.458,MN18080620,15.508,8.802,4.146,209.213
MN19060906,57.623,18.657,10.309,2374.089,MN19050624,21.246,14.312,4.924,313.469
MN18020111,50.28,18.706,7.477,1365.192,MN18090106,34.485,19.419,9.047,1174.43
MN18020922,21.323,8.338,4.765,176.249,MN19050620,28.074,7.7,5.343,390.702
MN19061411,30.276,9.927,5.125,598.296,MN19050629,18.659,10.688,4.315,264.399
MN19061109,0.142,0.085,0.057,0,MN18080436,12.781,6.02,4.186,79.845
MN19061004,33.186,17.832,7.414,1188.384,MN19081025,36.267,9.491,4.903,731.625
MN19081309,14.668,7.388,3.742,59.017,MN18080433,17.793,7.717,5.344,186.449
MN18020240,52.456,13.402,7.069,1019.448,MN19081038,24.512,10.492,3.692,262.486
MN18020239,32.235,11.108,7.179,762.525,MN11AA0411,44.271,8.917,4.934,461.43
MN19061003,33.382,14.339,7.288,938.256,MN11AA0705,51.143,13.344,9.457,1965.39
MN13AA0722,35.505,14.176,7.586,1049.145,MN19081024,32.979,11.28,7.306,976.593
MN19081411,30.123,2.812,1.155,10.758,MN18080431,25.551,8.027,3.668,211.045
MN18020243,33.703,21.962,7.594,1445.019,MN19050625,15.925,12.077,5.467,275.533
MN19061218,18.679,11.417,4.504,215.092,MN18080514,28.328,13.492,5.224,454.163
MN13AA0721,27.296,18.03,6.287,629.836,MN18090213,33.868,10.053,5.295,431.905
MN14AA0410,24.901,8.993,6.548,282.316,MN18080704,44.191,14.902,10.113,1667.504
MN18020921,11.665,9.506,3.908,116.52,MN19080942,32.655,8.167,5.932,415.33
MN19081507,13.928,10.055,5.423,202.165,MN19080844,28.938,12.171,10.558,1079.39
MN14AA0329,25.972,7.569,4.376,172.901,MN19080941,33.606,16.216,6.372,775.18
MN14AA0423,18.873,7.243,3.917,161.762,MN19050616,26.451,16.825,7.907,945.926
MN19062007,20.914,9.536,4.105,298.867,MN19081030,28.22,12.528,6.4,781.137
MN19062106,36.08,19.239,7.597,1223.51,MN19081031,23.172,11.036,6.437,555.471
MN19081504,26.76,5.437,2.195,67.471,MN11AA0608,34.768,12.586,9.094,812.88
MN18020241,47.517,6.657,4.123,299.636,MN18090310,29.048,8.222,4.863,290.49
MN18020832,21.785,10.128,6.752,336.407,MN19080937,33.887,11.812,6.422,827.455
MN19061606,32.946,9.77,7.468,644.028,MN19050609,35.304,17.295,8.763,1729.259
MN19061110,0.141,0.081,0.055,0,MN11AA0412,35.1,9.14,5.914,556.713
MN14AA0320,14.382,7.193,1.626,54.439,MN19081028,28.618,16.962,5.758,700.576
MN19062108,32.202,7.736,2.718,145.688,MN18080619,20.882,10.303,4.079,370.864
MN19062003,54.835,20.097,10.872,2985.124,MN19050255,26.238,11.429,5.714,435.136
MN10AA1004,59.735,20.639,11.01,5061.162,MN18080427,29.055,8.141,6.581,446.215
MN12AA1108,66.208,20.054,6.827,3301.157,MN19050550,37.557,12.243,6.998,987.42
MN12AA5604,68.444,21.544,9.315,3394.424,MN19050257,30.313,9.166,5.015,236.542
MN12AA0507,66.159,18.469,9.616,3106.812,MN19050618,23.161,15.865,5.591,442.498
MN12AA6116,57.33,6.208,4.645,443.257,MN19081036,23.142,9.553,3.052,159.369
MN19061206,59.094,21.299,10.085,2683.975,MN19050613,33.491,14.954,6.42,758.073
MN13AA0211,55.55,20.641,9.802,2718.608,MN19081009,34.888,15.51,5.804,523.496
MN13AA0313,54.799,17.889,6.391,1364.069,MN18090107,27.338,11.207,7.035,380.821
MN12AA1109,69.387,16.723,7.528,2305.565,MN18080214,26.646,13.734,5.628,638.114
MN12AA6608,72.569,14.787,6.808,1471.802,MN19050619,36.797,8.421,5.579,457.008
MN12AA6606,71.562,22.032,8.714,4085.483,MN18080420,28.982,20.504,8.696,990.355
MN12AA6407,46.971,19.871,4.628,895.103,MN19081034,24.033,8.478,7.756,591.18
MN12AA0207,55.286,22.124,7.779,3689.891,MN18080516,33.106,10.989,4.896,416.5
MN12AA0302,54.831,14.857,7.997,2170.832,MN18080421,40.119,14.405,5.413,1095.46
MN10AA0905,60.143,14.233,9.732,2619.489,MN19050534,36.595,15.216,14.606,2129.891
MN12AA6615,57.713,9.075,4.189,571.043,MN19050615,25.779,23.646,6.498,998.443
MN13AA0111,58.131,21.147,7.27,2935.424,MN19050814,8.854,2.836,2.308,3.234
MN12AA6110,52.977,12.69,9.46,1751.417,MN18080307,32.13,13.945,5.266,731.099
MN12AA1110,48.533,23.757,7.328,2032.176,MN19050809,33.292,17.415,8.137,1479.269
MN19061304,64.942,15.408,10.851,2820.604,MN19080823,47.622,13.098,6.432,808.955
MN12AA6610,42.13,18.226,4.644,1321.262,MN13AA0619,27.452,9.077,4.381,246.93
MN10AA0706,67.862,13.386,8.656,2807.132,MN13AA1123,26.393,12.486,8.43,1117.144
MN19061305,66.285,10.394,6.318,1093.082,MN19080923,60.42,15.523,6.674,2663.055
MN12AA6409,44.779,12.743,7.379,827.519,MN11AA0811,32.465,15.916,8.197,979.829
MN12AA6514,35.113,7.51,4.079,249.18,MN19080604,24.115,7.659,2.546,219.42
MN12AA6008,54.224,18.043,8.058,2620.859,MN19050409,26.359,12.614,5.48,418.385
MN12AA6509,58.648,20.111,8.193,1439.709,MN13AA1217,25.887,12.83,4.257,340.898
MN12AA6013,54.739,5.745,3.838,320.419,MN13AA0617,23.019,8.935,3.214,154.797
MN12AA6009,58.584,20.075,7.055,2045.758,MN19065024,12.707,6.633,3.338,101.96
MN10AA0107,40.241,11.512,5.123,656.894,MN19080817,38.58,18.276,9.806,1968.672
MN12AA0410,36.196,12.744,5.768,733.911,MN13AA1121,27.062,11.023,8.133,879.04
MN12AA6216,22.737,6.058,4.528,196.592,MN19063704,62.835,17.361,4.625,1576.142
MN12AA1113,33.938,22.327,7.001,1398.961,MN19081018,49.064,10.066,8.142,1854.567
MN10AA0207,40.1,19.093,10.227,2723.421,MN13AA1029,32.432,15.012,9.038,1259.924
MN19063005,36.298,15.357,8.397,1035.367,MN18080213,34.417,12.605,6.601,360.938
MN19060605,42.796,19.181,8.208,1586.893,MN18060709,61.878,20.239,9.54,3802.655
MN19063007,25.101,9.653,3.634,236.443,MN13AA1031,25.937,15.264,10.158,858.743
MN12AA6017,38.611,18.249,7.469,1294.809,MN13AA0510,24.024,14.951,6.173,297.221
MN12AA6018,35.098,12.745,3.189,280.678,MN19081206,59.062,11.836,7.177,2119.4
MN12AA0216,34.722,13.109,7.545,915.844,MN13AA1128,31.261,10.256,4.657,241.651
MN13AA0116,24.878,18.783,10.039,1410.735,MN13AA0513,32.444,12.5,8.385,738.11
MN10AA0809,33.596,20.283,8.922,1561.87,MN19064812,14.099,7.196,2.777,70.033
MN10AA1005,50.108,19.116,8.982,3915.446,MN19065009,53.19,17.549,8.918,2109.207
MN10AA1006,47.196,8.179,5.952,920.165,MN19080843,42.045,11.283,6.487,962.047
MN10AA0908,47.234,15.839,9.364,1957.713,MN19063904,60.386,13.324,7.811,2091.972
MN12AA6105,44.488,13.327,6.291,533.381,MN13AA1219,30.164,13.781,5.034,714.214
MN10AA0607,43.535,13.435,10.406,2156.594,MN13AA1120,30.851,16.671,6.352,880.583
MN12AA0209,33.549,21.482,8.075,1336.389,MN19063811,19.578,7.974,4.068,156.862
MN19063107,40.833,13.689,5.146,888.362,MN19080846,41.447,10.449,4.507,479.173
MN10AA0910,38.113,14.631,6.887,917.613,MN13AA0816,17.826,11.492,5.382,252.195
MN10AA0108,38.932,10.601,5.031,395.335,MN13AA0512,28.937,16.853,13.642,0
MN12AA6517,41.014,8.187,3.798,254.742,MN11AA0606,42.68,19.258,10.744,2880.474
MN12AA6618,33.962,13.133,5.441,671.721,MN19065109,22.344,9.818,4.415,222.023
MN13AA0216,35.687,12.417,4.952,673.818,MN18080612,34.752,6.875,5.19,358.124
MN12AA0807,41.104,23.98,9.642,2075.364,MN19063603,56.47,21.481,10.175,2392.427
MN12AA0511,43.131,19.312,5.387,1563.412,MN19064207,67.612,12.722,5.323,1165.208
MN12AA6109,36.216,14.662,6.32,874.553,MN13AA1125,28.859,14.307,7.711,807.769
MN10AA1011,36.949,14.005,7.074,1358.555,MN19064506,46.997,6.976,5.553,503.879
MN12AA0509,45.745,20.123,9.576,2323.268,MN13AA0512,24.259,11.91,6.738,380.657
MN19062906,42.3,17.307,9.24,1315.898,MN13AA1220,33.371,8.141,6.337,553.031
MN19062904,43.857,19.096,10.218,1423.917,MN19080842,39.973,8.682,5.31,764.447
MN12AA0512,47.651,20.296,8.525,2495.179,MN13AA1028,37.046,14.815,7.421,1015.085
MN10AA0710,44.462,11.699,7.424,1423.768,MN18080208,50.192,23.19,6.362,2356.042
MN13AA0215,31.72,12.084,5.173,463.429,MN19064505,54.459,10.868,7.541,1050.31
MN12AA0808,35.456,21.489,9.414,1998.097,MN19080826,45.236,9.787,8.778,1224.136
MN19061306,49.822,16.638,8.274,1708.451,MN13AA1124,27.347,9.223,8.064,654.263
MN12AA6106,34.519,14.683,7.382,1163.212,MN11AA0810,42.707,12.176,10.966,1067.109
MN12AA1115,28.029,11.223,8.48,278.614,MN18090808,42.655,11.15,5.599,785.776
MN13AA0319,37.958,8.834,3.421,144.521,MN13AA0416,27.368,16.552,6.281,411.313
MN10AA1007,41.684,17.961,11.133,2429.325,MN18080211,36.113,15.688,9.878,1666.108
MN12AA6217,34.724,10.899,3.854,301.816,MN13AA1032,23.405,11.377,7.061,511.57
MN10AA2011,31.644,13.958,4.86,384.59,MN19080921,67.552,14.654,7.591,1710.053
MN10AA0105,44.467,21.58,11.008,2043.746,MN19064805,52.934,10.377,8.714,914.166
MN10AA0405,39.026,10.257,6.449,502.454,MN19064404,49.521,15.939,6.668,1578.493
MN12AA6611,43.133,16.111,3.956,620.082,MN13AA1216,29.557,12.123,5.879,462.043
MN13AA0332,38.255,9.355,5.08,585.84,MN19065307,59.303,13.026,9.329,1546.183
MN13AA0115,31.75,12.145,3.174,281.984,MN13AA1035,29.822,13.501,4.991,320.262
MN10AA0413,39.673,12.068,8.334,1262.645,MN19080821,42.935,15.82,10.044,1557.215
MN19061803,50.291,19.803,10.688,2879.58,MN13AA0415,34.607,10.664,4.391,759.873
MN13AA0118,30.179,18.851,9.244,1168.022,MN19063906,55.285,14.376,6.768,1658.351
MN12AA6617,28.748,12.237,8.585,771.132,MN13AA0915,29.572,14.459,5.079,446.584
MN12AA0704,44.211,20.205,11.262,1875.316,MN19065306,57.19,13.937,7.042,1571.402
MN10AA2009,39.732,16.004,4.64,663.941,MN18090809,36.611,15.302,4.737,655.121
MN10AA1804,33.74,11.091,7.334,445.333,MN19081109,53.206,14.573,9.45,2231.016
MN12AA6515,46.119,6.875,3.539,288.499,MN19080924,64.569,13.849,7.417,1862.921
MN13AA0316,40.465,8.046,3.877,290.629,MN18080610,54.987,15.448,5.854,1571.44
MN12AA6616,36.285,10.899,7.104,766.716,MN19065103,62.739,21.28,12.384,2391.944
MN13AA0114,34.648,13.111,6.344,511.027,MN11AA0809,35.191,11.56,6.221,668.531
MN12AA0910,32.108,18.726,8.659,1915.574,MN19080702,60.791,17.717,9.785,2032.692
MN10AA0106,43.611,9.634,7.282,815.479,MN18090810,53.229,9.758,3.72,668.757
MN10AA0509,36.12,10.841,6.062,559.09,MN18090806,53.582,14.441,8.085,1471.215
MN19061308,37.571,9.511,5.05,408.633,MN13AA0615,33.276,11.476,4.187,462.903
MN12AA6609,40.213,15.769,3.767,871.166,MN19065304,66.907,22.691,11.046,4026.085
MN10AA0909,43.055,13.769,5.574,953.39,MN13AA0420,30.111,7.699,2.53,171.469
MN10AA0109,37.045,8.859,5.578,462.208,MN13AA1116,24.934,14.391,5.681,483.843
MN12AA6218,21.934,7.782,3.135,113.533,MN19080814,54.076,13.662,12.238,2559.666
MN12AA6016,36.908,21.159,5.932,767.717,MN19080910,69.044,12.325,10.844,2566.378
MN12AA5605,44.729,15.242,10.88,883.904,MN19080819,33.047,16.154,9.297,1423.056
MN12AA0217,40.992,14.714,7.983,1502.796,MN18080212,47.329,8.949,6.183,911.295
MN12AA6613,39.163,15.798,3.797,571.373,MN13AA1030,30.791,10.754,5.486,224.226
MN19060609,35.101,13.607,5.164,454.942,MN19080822,42.142,15.278,8.801,1804.36
MN12AA6112,49.95,10.51,7.246,715.535,MN19080815,51.681,12.646,7.221,1663.942
MN12AA0912,46.437,22.876,10.611,4140.496,MN18080615,31.425,10.843,5.28,519.72
MN12AA5620,34.309,11.761,6.916,516.811,MN13AA0418,21.554,9.333,3.867,174.393
MN12AA6512,30.907,12.662,5.328,821.973,MN13AA0814,22.436,11.923,4.8,304.368
MN12AA6612,41.802,10.14,5.641,477.437,MN13AA1027,30.05,15.804,6.446,765.957
MN12AA6011,35.746,20.282,7.8,1726.47,MN19064811,18.743,7.37,3.143,103.296
MN19061104,46.623,22.027,11.044,2372.247,MN19063813,21.974,5.74,3.214,110.387
MN12AA0305,29.686,10.947,2.65,327.859,MN19080841,27.939,14.551,8.774,950.484
MN12AA0705,40.76,9.235,4.666,578.636,MN18080613,30.225,9.527,6.148,544.423
MN13AA0119,35.812,17.057,10.135,1504.942,MN19065011,48.905,22.754,5.647,1506.764
MN19063006,46.25,15.726,5.378,785.572,MN19080847,30.824,11.108,5.656,556.458
MN10AA0708,51.436,19.831,8.694,2477.636,MN18080210,52.086,15.787,8.403,1764.719
MN19060606,41.755,21.059,8.193,1237.197,MN19063812,20.766,7.244,4.452,183.763
MN12AA6113a,32.637,18.817,6.509,1019.624,MN13AA1114,29.736,6.151,5.83,372.759
MN12AA6107,28.485,12.424,6.489,574.531,MN19080603,22.073,7.545,7.253,407.524
MN10AA0808,23.741,12.04,9.092,758.698,MN19081208,59.584,15.866,10.251,2963.049
MN19061806,24.534,13.884,6.687,468.725,MN18090707,42.886,12.174,4.067,617.234
MN10AA0408a,27.367,13.745,7.172,571.909,MN19063804,59.465,18.072,9.542,2592.018
MN12AA0119,25.729,15.443,4.143,373.077,MN13AA1117,28.534,12.414,7.175,839.504
MN10AA0511,21.808,9.817,4.386,258.202,MN19080816,54.122,13.969,8.615,1458.387
MN10AA0407,27.113,14.135,7.261,760.853,MN18032106,19.091,11.639,7.104,457.36
MN10AA0410,18.163,13.843,4.592,352.678,MN18101427,19.109,11.313,4.494,322.843
MN12AA5709,36.188,16.374,5.908,990.712,MN13AA1115,34.934,8.449,3.907,248.932
MN12AA0113,22.426,18.586,5.904,667.84,MN18071719,17.802,9.368,4.066,189.04
MN12AA0118,24.242,13.335,3.687,272.757,MN18071819,35.485,12.97,10.847,1722.623
MN12AA1117,30.304,17.196,5.078,512.674,MN18071119,30.408,18.979,6.224,552.175
MN10AA0912,21.316,12.915,8.547,728.08,MN13AA1020,42.8,20.518,7.308,1866.047
MN12AA6119,22.957,12.292,5.204,261.92,MN18101428,22.17,11.304,3.854,230.212
MN19063113,11.968,9.984,4.195,147.158,MN13AA0511,42.125,15.604,6.471,1067.419
MN19061310,21.822,9.655,5.883,346.939,MN18031226,21.068,14.397,5.115,434.631
MN12AA5710,20.054,16.33,4.777,456.248,MN18030421,21.538,12.006,6.168,478.353
MN19061307,21.703,11.003,5.271,525.883,MN18072336,23.487,16.923,6.399,785.901
MN12AA0122,20.676,14.178,4.463,287.656,MN18032006,35.351,20.697,7.046,1619.985
MN10AA0409,21.009,8.645,6.296,399.812,MN18071418,22.809,15.365,5.144,443.092
MN10AA0110,30.911,11.753,5.176,508.664,MN13AA1022,46.325,18.816,9.592,2751.21
MN10AA0612,30.204,12.634,7.486,633.651,MN18101443,19.884,6.658,4.162,152.039
MN10AA1908,31.732,15.378,5.779,496.682,MN18072335,23.143,15.318,6.028,614.105
MN12AA0111,34.177,20.263,4.277,813.005,MN18072020,34.183,14.066,7.898,1411.458
MN12AA0110,24.072,13.872,5.266,369.822,MN18031920,29.086,13.989,5.485,472.497
MN19063118,16.253,6.982,3.46,93.06,MN18071417,22.55,16.596,7.704,1126.367
MN10AA1014,26.72,17.305,5.642,608.992,MN18101420,32.66,12.074,5.653,520.517
MN12AA0115,28.8,12.049,6.554,660.071,MN18101431,15.73,8.586,5.759,285.463
MN12AA5613,23.414,7.643,4.183,181.467,MN18071620,29.025,16.042,4.582,567.049
MN19063111,30.047,9.09,4.578,308.299,MN18071617,33.503,18.881,3.936,1055.168
MN12AA0920,28.749,15.56,5.735,597.044,MN18101437,15.889,8.294,4.594,210.403
MN19060611,15.325,10.502,4.121,183.049,MN18030420,21.511,11.777,11.092,673.987
MN10AA0913,14.402,11.355,4.518,230.134,MN13AA0616,48.509,12.469,6.315,636.203
MN10AA0408b,30.349,21.573,8.269,1129.123,MN18031218,38.818,13.12,9.203,1266.539
MN19063012,21.381,7.396,5.429,148.893,MN18072334,32.029,14.598,8.243,744.467
MN12AA6518,24.26,10.617,3.82,229.439,MN13AA1122,36.953,17.435,9.047,1163.619
MN19063112,23.73,6.497,4.087,253.036,MN18071717,20.1,8.499,3.695,173.358
MN12AA0916,32.976,24.763,11.629,2487.133,MN13AA1109,49.339,10.615,6.603,441.166
MN12AA5814,30.401,13.405,7.283,827.528,MN18031013,20.982,10.908,5.147,295.373
MN12AA0909,33.629,13.629,7.159,1035.32,MN18031919,28.062,17.572,7.932,779.312
MN19063110,20.191,11.899,4.874,272.153,MN18071822,31.941,14.554,7.423,934.306
MN10AA0111,22.752,8.388,4.874,202.627,MN18072407,31.921,16.754,5.277,705.074
MN10AA0114,22.653,10.385,6.295,312.658,MN18072104,31.727,10.776,5.836,474.079
MN10AA1806,24.022,12.863,4.493,373.913,MN13AA1129,44.949,11.724,6.056,786.116
MN12AA0116,29.504,9.152,4.442,224.653,MN18031114,48.161,5.861,3.277,279.52
MN12AA0112,29.743,15.297,8.597,1674.931,MN18071815,36.899,17.861,7.554,1824.565
MN12AA0121,20.412,15.486,6.324,630.986,MN18031008,24.038,15.688,7.457,976.602
MN12AA0409,27.816,14.497,9.202,984.51,MN18071923,34.001,17.803,7.454,1688.794
MN10AA2008,35.484,14.011,7.93,863.009,MN18101432,25.884,8.319,4.988,196.494
MN19061309,21.719,9.327,6.047,382.195,MN18101424,18.182,14.73,4.629,531.096
MN19063117,16.799,6.83,5.034,155.178,MN18032010,21.295,13.066,8.046,592.477
MN19062404,35.142,14.261,5.78,921.879,MN18071420,26.245,15.015,4.544,245.396
MN12AA6113b,30.241,15.424,4.023,525.572,MN18072716,26.534,11.466,5.234,467.427
MN12AA5815,23.057,15.11,5.894,514.29,MN13AA1021,39.118,24.109,7.19,2288.688
MN12AA0907,34.803,18.388,8.031,1640.972,MN18072323,36.591,17.579,7.644,1237.142
MN12AA0114,26.14,14.651,10.49,940.896,MN18031216,33.945,20.81,6.452,1280.234
MN12AA0214,28.343,12.47,7.323,704.505,MN18072103,33.013,17.147,5.239,681.765
MN10AA0208,32.434,11.464,5.72,498.359,MN18071703,23.92,17.703,10.137,1579.215
MN12AA6014,31.562,17.117,7.028,1108.087,MN18031217,48.374,12.237,7.498,1649.01
MN12AA0921,23.232,14.841,5.744,512.064,MN18032009,31.357,13.362,7.611,1028.809
MN10AA0609,24.735,13.552,10.441,758.496,MN18101423,26.929,13.784,4.911,317.246
MN12AA5810,17.981,13.647,6.007,262.57,MN18031916,30.438,17.479,9.434,1018.327
MN12AA0611,25.212,12.37,5.804,529.503,MN18032111,23.085,8.756,3.603,218.605
MN12AA6516,26.423,13.94,3.326,240.462,MN18101422,28.076,12.585,5.21,539.887
MN12AA6213,24.312,11.557,7.325,685.126,MN18031318,21.33,9.112,3.625,166.746
MN19061805,24.775,15.326,5.652,739.629,MN13AA0912,48.284,14.407,5.818,1184.635
MN10AA0811,31.536,7.14,3.722,330.044,MN18072337,21.832,14.112,6.728,554.618
MN10AA0116,16.645,11.898,5.301,342.241,MN18072331,23.363,18.637,7.392,1000.919
MN19063114,15.522,9.658,4.791,152.827,MN18031313,27.371,14.211,7.112,794.108
MN12AA5618,18.028,6.062,2.404,77.975,MN18071824,29.478,9.224,9.17,701.686
MN10AA0507,15.479,13.283,6.265,398.253,MN18072330,24.661,19.363,6.61,904.083
MN10AA0317,23.864,5.937,4.464,235.718,MN18101435,26.051,7.93,4.019,156.437
MN12AA0117,26.345,11.933,8.937,462.829,MN13AA0811,43.527,15.29,5.686,868.641
MN12AA5615,26.272,7.399,3.405,166.013,MN13AA1019,38.022,17.96,8.277,1756.653
MN12AA0919,22.86,17.451,10.758,947.922,MN13AA1118,46.792,6.233,3.998,285.192
MN12AA5614,17.528,7.606,3.003,170.465,MN13AA1016,46.01,14.278,6.962,1439.383
MN12AA0915,27.448,23.421,9.452,1882.444,MN18031118,19.234,14.27,5.651,378.936
MN12AA6015,34.595,17.562,5.135,638.828,MN13AA1112,47.946,10.388,8.336,1201.244
MN10AA1008,30.003,9.491,6.697,490.705,MN13AA1126,40.094,18.601,5.692,1017.777
MN10AA1805,19.489,14.823,4.693,416.453,MN18031122,23.571,8.595,2.537,153.816
MN19061804,27.362,15.579,6.029,832.246,MN18031220,37.869,10.425,10.326,1090.691
MN12AA6511,26.234,9.249,6.175,352.032,MN18031010,35.956,11.027,4.066,449.146
MN12AA5612,15.101,11.072,3.675,194.788,MN13AA1018,41.911,16.909,8.735,1579.148
MN10AA0209,27.261,10.899,5.155,400.246,MN18031317,25.452,8.718,6.029,320.213
MN10AA0712,19.625,9.874,6.136,380.176,MN18031009,33.422,13.839,7.274,783.333
MN19063109,22.346,12.387,6.362,371.91,MN13AA0809,53.769,15.834,7.608,2375.836
MN10AA0711,30.66,12.91,4.818,519.736,MN18071716,25.454,7.571,4.587,279.188
MN12AA0215,32.937,12.442,7.508,720.671,MN18032104,23.834,17.681,9.198,1044.654
MN12AA6314,28.562,15.839,6.502,699.964,MN18071120,31.055,15.882,7.004,662.885
MN10AA1010,16.239,11.023,6.559,295.66,MN18032108,28.633,5.826,2.883,151.117
MN12AA0411,31.737,12.879,6.622,758.229,MN18101444,20.979,6.909,2.226,107.891
MN19063106,28.63,11.598,3.935,232.694,MN18032008,29.842,18.303,9.157,1765.071
MN12AA5617,23.727,15.467,9.06,966.545,MN13AA1214,41.108,12.095,8.19,1257.778
MN10AA0610,27.415,8.145,3.991,240.039,MN18031225,32.056,10.124,9.354,791.267
MN12AA1116,24.842,14.14,9.774,666.935,MN18031006,53.135,18.687,10.577,2619.149
MN12AA0123,21.224,12.825,9.007,699.195,MN18101434,23.77,6.69,6.32,329.657
MN12AA5811,29.844,15.183,7.529,641.757,MN18071921,35.788,14.466,7.435,1317.977
MN10AA0210,25.283,6.629,3.99,194.405,MN18031312,42.503,8.494,5.833,446.903
MN10AA2010,30.365,8.36,4.733,299.059,MN18072332,26.565,15.561,7.192,826.387
MN19063115,21.986,5.813,3.412,102.342,MN18101426,22.616,11.717,5.996,533.429
MN10AA0313,14.471,8.099,6.466,229.285,MN18030422,18.475,12.775,9.409,397.766
MN10AA0807,29.878,14.254,10.434,957.62,MN18101418,33.739,11.413,8.674,1038.295
MN12AA0210,27.036,22.196,8.254,1519.145,MN13AA1218,40.041,12.752,7.654,679.413
MN12AA0303,35.214,15.218,6.079,986.702,MN18101430,16.282,11.1,4.275,188.732
MN12AA0917,28.296,15.514,7.499,724.889,MN18031113,41.429,13.491,8.463,1240.607
MN19063008,23.561,12.359,3.818,211.276,MN18031015,15.344,8.787,3.844,142.684
MN12AA6012,32.034,10.51,8.326,684.972,MN18101425,19.321,13.123,3.675,380.996
MN19063116,17.822,8.42,5.514,200.92,MN18071707,18.506,14.904,4.312,354.33
MN10AA1012,24.226,18.945,10.223,1440.861,MN18071828,25.017,9.525,7.47,452.388
MN19063108,29.107,14.171,4.784,381.194,MN18071622,25.679,18.41,8.625,737.625
MN12AA0304,27.347,15.902,6.204,587.8,MN18031221,51.256,10.68,6.406,1177.777
MN12AA6410,23.966,14.126,6.653,553.39,MN18032105,18.646,12.344,7.685,501.346
MN19063009,19.724,8.826,4.637,273.877,MN18072023,33.455,15.113,6.534,976.879
MN10AA1009,25.006,14.569,5.227,539.47,MN18071709,24.16,13.093,5.127,315.927
MN10AA0812,27.329,10.812,7.585,544.398,MN18031316,27.966,11.364,5.61,386.3
MN10AA0115,29.8,9.549,5.438,320.717,MN18071706,25.926,11.211,7.48,450.201
MN19060607,20.352,19.186,8.302,839.107,MN18031314,21.386,15.805,4.803,601.253
MN12AA0514,28.186,16.629,6.531,954.772,MN18072327,28.36,22.839,8.492,1255.46
MN12AA6513,26.291,12.576,6.746,619.991,MN18072324,34.355,18.426,5.016,947.673
MN12AA6108,26.337,16.828,7.498,1073.584,MN18032102,23.117,21.106,9.182,1232.031
MN12AA5813,23.008,13.499,7.619,454.764,MN18031227,21.715,7.141,4.22,217.969
MN12AA0515,17.236,14.163,4.056,304.083,MN18101429,16.773,10.35,6.993,260.86
MN10AA0911,31.235,11.944,6.9,591.738,MN18072325,36.456,20.618,5.507,1376.638
MN12AA6510,23.801,16.817,5.383,665.613,MN18031315,25.893,9.448,4.803,320.537
MN10AA0311,26.436,13.289,5.03,466.562,MN13AA1017,48.567,14.447,7.656,936.901
MN12AA0918,31.718,13.976,7.642,784.166,MN18101417,36.716,14.062,9.213,1107.786
MN19060610,21.765,9.927,3.98,279.657,MN18031219,29.303,21.366,6.857,1130.555
MN19063011,16.04,10.469,5.608,243.31,MN18071122,27.49,15.458,2.527,285.361
MN10AA0608,35.917,9.397,5.5,667.07,MN18032103,23.218,20.43,6.722,787.223
MN12AA6111,21.004,19.344,7.881,998.377,MN18032110,21.443,10.156,5.515,384.499
MN10AA0406,24.954,9.982,7.1,539.783,MN18101433,25.897,7.402,3.474,218.235
MN19060608,22.387,11.139,5.624,415.31,MN18031117,24.614,10.533,4.947,328.555
MN12AA5616,32.194,16.324,6.249,800.54,MN13AA0916,40.299,12.751,7.98,935.345
MN10AA0510,31.338,9.043,4.383,370.775,MN18072338,18.43,10.635,3.293,154.013
MN19064211,44.222,10.019,5.132,764.285,MN18031228,15.899,13.577,3.965,198.155
MN18060721,32.039,15.515,4.791,411.269,MN18032109,18.41,14.151,6.371,353.231
MN19063907,45.25,12.768,9.485,704.769,MN13AA0914,44.094,15.155,7.723,1159.545
MN18061013,25.637,14.89,7.658,771.184,MN18071616,35.664,19.918,10.324,2142.372
MN19080602,22.936,16.227,6.269,532.134,MN18031311,51.012,10.373,6.08,744.401
MN19080605,18.093,15.76,4.458,266.76,MN18101312,30.018,11.185,5.837,454.821
MN19063707,16.175,10.47,5.643,304.938,MN18101436,25.747,10.13,4.972,136.763
MN18061210,35.504,8.921,5.884,497.017,MN18101421,32.134,10.893,3.193,232.477
MN18060724,30.641,10.877,4.962,523.372,MN18031115,37.364,11.745,4.865,588.255
MN18060712,47.872,18.602,8.72,1542.807,MN18072713,45.582,7.089,4.138,399.65
MN18061112,26.72,18.141,6.441,847.411,MN18110718,28.988,12.077,7.293,659.916
MN18061312,24.221,14.899,5.11,502.528,MN18101310,27.731,10.4,5.248,314.071
MN18060719,24.647,16.413,5.477,475.871,MN18072333,34.917,10.386,6.243,589.971
MN19065013,39.325,19.28,5.094,941.946,MN18110231,30.673,16.24,8.253,1238.455
MN18061110,21.57,14,5.554,400.948,MN18072714,34.237,6.715,5.614,497.032
MN18060808,35.563,16.643,7.468,1112.851,MN18110714,32.997,15.793,6.965,968.772
MN18060731,20.196,6.585,2.424,95.928,MN18072717,28.257,6.994,4.378,192.011
MN19063708,13.392,10.21,5.261,196.073,MN18072408,37.952,11.759,7.742,696.915
MN18061211,26.269,13.814,6.164,679.657,MN18071917,45.805,19.263,6.631,1375.059
MN18061016,20.099,13.108,5.612,331.972,MN18100206,30.124,16.312,8.682,790.601
MN19064109,25.705,12.642,5.27,370.173,MN18110235,33.987,15.718,7.848,1102.612
MN19063905,42.665,20.147,10.505,1631.654,MN18071821,40.064,10.222,3.716,586.429
MN19064409,16.111,9.99,5.057,203.697,MN18110239,30.781,11.958,5.202,620.498
MN18061613,23.206,13.06,5.496,370.377,MN18110237,31.924,21.042,3.924,640.565
MN18061308,29.35,15.55,5.804,677.627,MN18072014,46.19,22.09,9.115,2271.656
MN18061416,16.364,11.159,8.553,496.792,MN18072021,37.137,15.015,10.606,1140.363
MN18060720,32.776,13.913,5.543,794.299,MN18110712,30.911,21.065,7.407,1469.755
MN19064008,15.133,5.181,2.947,54.141,MN18101410,53.182,19.027,10.019,3250.483
MN18061310,30.463,17.457,4.433,461.896,MN18072317,43.213,17.616,11.211,1987.661
MN19065310,35.924,13.795,8.14,985.281,MN18101414,47.913,14.95,6.656,1738.466
MN19065309,31.236,15.557,7.561,1094.153,MN18110217,44.107,18.45,8.022,2827.937
MN19063609,16.527,13.002,3.683,155.55,MN18032007,52.234,13.747,7.022,1398.218
MN19064106,36.943,13.402,8.039,897.63,MN18072409,35.605,8.344,5.171,373.001
MN18061111,22.241,14.693,5.751,463.868,MN18071825,30.933,9.133,6.548,431.886
MN19063808,21.738,9.056,8.042,306.074,MN18072025,30.484,11.822,6.619,813.854
MN19065023,16.577,6.887,5.158,194.368,MN18110222,52.64,12.415,8.638,1651.083
MN19065022,17.922,6.18,4.951,144.167,MN18101412,48.583,18.725,10.736,1691.711
MN18061215,33.789,8.211,7.072,288.987,MN18110221,38.264,18.696,5.683,1417.023
MN18061415,29.126,10.17,5.677,354.717,MN18110227,53.309,15.43,9.631,1784.231
MN19064308,25.527,9.12,4.574,251.172,MN18072321,37.675,18.51,7.48,2012.825
MN19063805,41.334,17.323,8.11,1303.353,MN18031917,41.44,12.457,5.513,796.411
MN19064808,22.567,10.039,4.168,206.918,MN18071116,33.979,20.027,5.958,1094.604
MN19063909,34.652,6.186,2.883,156.634,MN18072611,45.816,17.463,8.821,1636.847
MN19065021,23.334,7.209,3.619,112.719,MN18072715,35.397,4.859,3.967,173.971
MN18061011,34.656,17.96,10.157,1153.04,MN18110244,24.162,11.244,5.409,375.88
MN19064105,28.013,18.6,10.717,1018.46,MN18110620,31.664,17.243,6.603,991.385
MN18061214,18.547,10.716,7.26,515.489,MN18110240,22.756,12.884,7.234,603.252
MN19064005,17.267,11.481,3.19,155.882,MN18032107,32.99,9.012,3.79,252.125
MN18060714,35.316,15.403,7.587,967.393,MN18072406,31.507,13,5.717,822.414
MN18060725,32.423,10.204,7.052,501.591,MN18101316,20.703,11.661,5.915,210.61
MN19064809,12.904,9.424,4.369,208.061,MN18072320,42.094,22.713,7.491,1419.834
MN19064208,30.128,13.038,6.619,805.839,MN18071710,33.876,14.964,3.593,309.85
MN19063611,13.516,8.115,2.796,75.877,MN18072102,49.794,17.487,8.106,1935.836
MN18061414,26.149,10.965,7.81,649.357,MN18072318,42.68,19.987,10.1,1775.151
MN19064110,14.554,8.682,2.949,76.118,MN18071415,30.046,22.338,8.65,1954.488
MN19064408,26.824,7.839,3.614,161.333,MN18071820,39.101,10.397,5.369,672.311
MN19064806,19.624,10.45,4.097,265.042,MN18071918,53.172,18.563,8.007,2508.216
MN19064006,17.734,7.87,6.069,225.75,MN18072903,35.967,13.762,6.79,1168.003
MN19063606,26.555,10.033,4.003,429.747,MN18110243,27.035,7.856,7.562,565.308
MN19065019,24.237,8.081,3.868,231.401,MN18072040,40.302,12.831,8.2,1353.256
MN18061612,32.801,11.535,3.664,296.268,MN18110228,32.354,21.581,6.826,1713.047
MN18061109,31.25,10.755,5.933,416.946,MN18072316,51.954,18.336,10.419,1497.266
MN18061615,22.674,7.957,3.315,202.293,MN18072017,36.299,19.593,9.525,1431.743
MN19065017,18.618,14.923,4.602,274.101,MN18110505,19.842,10.556,8.27,337.864
MN18061209,21.451,15.078,10.326,903.37,MN18110623,25.693,13.606,5.553,546.279
MN18061313,22.922,14.678,5.386,451.232,MN18031918,33.845,15.089,7.501,729.051
MN19064912,24.768,10.685,4.979,283.661,MN18071123,20.746,9.993,6.007,270.539
MN19064007,16.664,6.948,3.137,74.684,MN18110108,42.852,19.516,11.245,1315.358
MN18061212a,21.46,14.134,6.857,491.088,MN18031921,27.258,9.818,5.092,242.844
MN18061418,23.915,7.194,4.523,242.424,MN18072502,42.33,20.008,11.113,2483.398
MN19064306,26.018,12.317,8.426,570.744,MN18101311,19.388,13.385,5.063,393.766
MN18061208,29.369,18.669,5.945,809.858,MN18110230,28.437,16.169,9.516,1329.433
MN19063504,30,8.923,4.738,397.667,MN18101309,22.949,15.429,5.561,654.445
MN19064407,22.454,7.453,5.553,317.307,MN18072901,41.542,20.149,8.446,1910.117
MN19063605,32.629,14.475,10.674,578.819,MN18072016,37.293,20.933,8.853,1278.503
MN19064104,47.381,18.851,9.887,1797.609,MN18071414,37.138,10.639,3.173,236.491
MN18060729,17.504,13.889,3.832,211.336,MN18031111,48.443,17.37,5.723,1224.138
MN18061107,26.44,16.089,6.113,581.628,MN18072503,33.28,15.898,7.132,1462.477
MN19063706,18.519,11.393,8.675,406.784,MN18072902,31.134,19.477,7.394,1758.917
MN18060718,31.169,14.812,5.509,520.796,MN18071625,28.185,9.95,4.461,408.87
MN18061614,26.957,11.204,5.073,349.039,MN18101307,35.735,17.318,6.364,1002.776
MN18061417,26.543,6.885,5.323,313.253,MN18072030,23.78,13.733,4.887,356.263
MN19064213,30.415,12.889,3.792,492.148,MN18101413,42.607,15.129,8.574,1557.876
MN18061212b,15.904,15.092,7.625,619.893,MN18072908,28.227,9.617,6.259,471.762
MN19064909,32.802,13.749,8.218,678.575,MN18071117,37.122,12.96,2.923,235.251
MN19063713,14.518,5.753,3.972,91.459,MN18072029,27.149,10.298,8.038,631.28
MN19064807,19.967,15.842,7.165,261.406,MN18032101,41.785,12.495,9.41,1226.985
MN19063807,28.87,11.138,6.002,472.699,MN18072319,50.294,15.048,6.48,1483.981
MN18061617,16.457,12.529,5.647,196.654,MN18110234,34.78,13.108,9.297,1053.538
MN19064108,30.148,7.932,4.315,292.365,MN18071701,44.436,21.362,8.208,1310.911
MN19065014,39.317,11.918,5.76,831.704,MN18110713,32.859,16.159,9.462,1661.619
MN19064405,28.336,8.049,5.259,353.783,MN18110238,31.712,12.74,7.434,796.164
MN18061309,29.326,13.5,4.758,465.994,MN18101317,19.851,9.58,4.345,223.267
MN19063403,20.158,10.061,5.081,279.943,MN18071922,40.775,16.752,6.802,522.031
MN18061014,25.031,14.485,7.094,555.017,MN18110241,27.855,12.271,3.88,393.662
MN18060734,15.689,8.735,3.354,126.78,MN18072018,43.45,12.991,9.086,1523.929
MN19060805,41.603,17.45,7.551,651.917,MN18101419,39.08,8.35,5.344,426.178
MN19065311,35.516,18.468,6.789,1031.649,MN18110622,28.852,7.593,6.193,501.866
MN19064403,34.561,14.041,6.352,832.092,MN18101308,24.784,20.969,6.967,795.367
MN18060807,35.037,9.706,8.321,706.338,MN18101313,23.681,10.651,4.698,189.161
MN19065312,29.048,15.415,10.009,896.685,MN18101415,43.282,13.35,5.446,1099.051
MN19064107,29.449,12.351,5.821,451.974,,,,,
\end{filecontents*}
\csvreader[column count=5, late after line=\\]{mesh_stats.csv}{}{\csvcoli & \csvcolii & \csvcoliii & \csvcoliv & \csvcolv & \csvcolvi & \csvcolvii & \csvcolviii & \csvcolix & \csvcolx}

\end{longtable}
\end{small}
\end{center}

\end{document}